\documentclass{article}

    \PassOptionsToPackage{numbers, compress}{natbib}

 \usepackage[main, final]{neurips_2026}

\usepackage[utf8]{inputenc} 
\usepackage[T1]{fontenc}    
\usepackage{hyperref}       
\usepackage{url}            
\usepackage{booktabs}       
\usepackage{amsfonts}       
\usepackage{nicefrac}       
\usepackage{microtype}      
\usepackage{xcolor}         
\usepackage{colortbl}

\newcommand{\tb}[1]{\textbf{#1}}
\newcommand{\ul}[1]{\underline{#1}}
\newcommand{\capa}{|\mathcal{A}|}

\definecolor{ramnet-red}{RGB}{198, 40, 40}
\definecolor{ramnet-green}{RGB}{46, 125, 50}
\definecolor{linkblue}{HTML}{248fd6}
\colorlet{ramrow}{linkblue!15}

\hypersetup{hidelinks}
\hypersetup{
  colorlinks=true,
  citecolor=linkblue,  
  linkcolor=linkblue,   
  urlcolor=linkblue     
}

\usepackage{placeins}
\usepackage{amsmath,bm}
\usepackage{amssymb}
\usepackage{graphicx}
\usepackage{caption}
\usepackage{subcaption} 
\usepackage{threeparttable}
\usepackage{multirow}
\usepackage[table]{xcolor}
\usepackage{footmisc}
\usepackage{wrapfig}
\usepackage[table]{xcolor}
\definecolor{highlight}{RGB}{255, 242, 230}

\title{RAM-Net: Linear-Time Sequence Modeling with Sparsely Addressable State}

\author{\textbf{Kaicheng Xiao}$^{\dagger}$ \quad \textbf{Haotian Li}$^{\dagger}$ \quad \textbf{Liran Dong} \quad \textbf{Guoliang Xing} \\
The Chinese University of Hong Kong \\
\texttt{\{xk023, lh023, dl123, glxing\}@ie.cuhk.edu.hk} \\[4pt]
\href{https://github.com/MuonCat/RAM-Net}{\textcolor{linkblue}{\nolinkurl{https://github.com/MuonCat/RAM-Net}}}
}

\begin{document}

\maketitle
{\renewcommand{\thefootnote}{$\dagger$}\footnotetext{Equal contribution.}}

\begin{abstract}
Linear attention offers an efficient alternative to full attention with a fixed-size recurrent state. However, this state is shared by all tokens, so information from distinct tokens becomes superposed within it and produces inter-token interference that degrades long-range fine-grained recall.
To address this issue, we propose RAM-Net, which replaces dense access to a shared state with sparse address-based access.
RAM-Net organizes the recurrent state as a fixed-size array of independent slots and uses an Address Decoder that maps each key or query into a sparse address, selecting a small subset of slots to write to or read from at each step.
This design directs tokens with non-overlapping addresses to disjoint slots, suppressing inter-token interference, while keeping per-step state access dependent only on the number of selected slots rather than the total state size.
Empirically, RAM-Net outperforms strong recurrent baselines on fine-grained long-range retrieval and achieves competitive perplexity and commonsense reasoning.
It does so while accessing fewer state elements per step than all baselines, e.g., $7.6\times$ fewer than Mamba2.
\end{abstract}

\section{Introduction}

Transformers~\cite{vaswani2017attention} have become the dominant architecture for sequence modeling, largely owing to the ability of full attention to capture long-range dependencies.
This ability rests on two design choices.
First, full attention stores every past token as a separate entry in a key-value (KV) cache, which keeps the information of each token separable.
Second, it weights these entries for each query through a softmax over query-key similarities, which tends to concentrate each read on the few entries that match the query.
However, such high-fidelity retrieval comes at a cost, as the KV cache grows linearly with sequence length and the total computation grows quadratically, creating a fundamental bottleneck for long-context modeling~\cite{pope2023efficiently}.

Linear attention removes this growth by compressing the history into a fixed-size recurrent state.
Early designs approximate the softmax similarity with kernel feature maps and accumulate each token into a single state matrix through an outer-product update~\cite{katharopoulos2020transformers, choromanski2020rethinking}.
Subsequent variants refine this update with data-dependent decay, gating, or delta-rule corrections~\cite{yang2024gla, qin2024hgrn2gatedlinearrnns, schlag2021linear, yang2024gated, dao2024transformers}.
Nevertheless, these designs give up both choices of full attention.
Tokens are superposed in one shared matrix rather than stored separately, and each step reads the entire state through linear aggregation rather than selecting a few entries.
Consequently, contributions from different tokens become entangled, and this inter-token interference degrades fine-grained long-range recall~\cite{arora2024zoology, sun2024learning}.
Moreover, since every step accesses the entire state, enlarging the state to hold more information raises the per-step cost proportionally.


However, we observe that a fixed-size state does not by itself require giving up separable storage or selective access.
This observation points to a third attention paradigm beyond full attention and linear attention.
In this paradigm, the state remains bounded but is divided into many separate units, and each step accesses only a few of them.
We realize this paradigm as the Random Access Memory Network (RAM-Net), which organizes the recurrent state as a fixed-size array of independent storage units called {\it slots}.
At each step, the key produces a {\it write address} that selects a few slots to write into, and the query produces a {\it read address} that selects a few slots to read from.
All other slots remain unchanged.
This address-based access resembles random access memory in computer systems and gives the architecture its name.
Since each address selects only a few slots, such sparse addressing brings two properties that answer the two problems of linear attention.
(i) \textcolor{linkblue}{\textbf{Confined interference.}} Tokens with non-overlapping addresses are written to disjoint slots, so interference arises only at the slots they jointly address.
(ii) \textcolor{linkblue}{\textbf{Decoupled state size.}} Each step accesses only the selected slots, so the total state can grow without increasing the per-step state access.


We evaluate RAM-Net at the 340M and 1.3B scales against Transformer++ and strong recurrent baselines.
On long-context retrieval, RAM-Net achieves the highest S-NIAH average at both scales.
On language modeling, it achieves the lowest WikiText perplexity at 340M and remains competitive at 1.3B on both perplexity and commonsense reasoning.
Notably, since sparse addressing accesses only a small subset of the state at each step, RAM-Net obtains these results with less per-step state access than every baseline, e.g., at least $7.6\times$ less than Mamba2 (Table~\ref{tab:main_results}).
This reduction also translates into practical speed, as RAM-Net Top-8 achieves $2.0\times$ the decoding throughput of the fastest baseline (Fig.~\ref{fig:tps}).

\section{Background}
\label{sec:background}
Attention enables a sequence model to associate past information with the current output through a state $\mathbf{S}_t$ that stores historical context.
At each step, two operators act on this state: a \textit{write} operator $\mathcal{W}$ that integrates the new input into $\mathbf{S}_t$, and a \textit{read} operator $\mathcal{R}$ that retrieves from $\mathbf{S}_t$ to produce the output~\citep{schlag2021linear, arora2024simple}.
Concretely, given an input row vector $\mathbf{x}_t \in \mathbb{R}^{1 \times d_m}$, the model computes a query $\mathbf{q}_t \in \mathbb{R}^{1 \times d_k}$,  key $\mathbf{k}_t \in \mathbb{R}^{1 \times d_k}$ and value $\mathbf{v}_t \in \mathbb{R}^{1 \times d_v}$ via learnable projections.
The key and value drive $\mathcal{W}$ to update the state, while the query drives $\mathcal{R}$ to retrieve from it:
\begin{equation}
\mathbf{S}_t = \mathcal{W}(\mathbf{S}_{t-1}, \mathbf{k}_t, \mathbf{v}_t),
\quad
\mathbf{o}_t = \mathcal{R}(\mathbf{S}_t, \mathbf{q}_t).
\end{equation}

Within this formulation, attention architectures differ in two design aspects.
(i) \textcolor{linkblue}{\textbf{State granularity}}: whether $\mathbf{S}_t$ is divided into separate units that each store part of the history or maintained as a single undivided block.
(ii) \textcolor{linkblue}{\textbf{Access pattern}}: how much of $\mathbf{S}_t$ the operators $\mathcal{W}$ and $\mathcal{R}$ access at each step, and how the accessed contents are weighted.
We next revisit full attention and linear attention along these two aspects.

\subsection{Full Attention}
\label{sec:full_attn}
\begin{figure}
    \centering
    \includegraphics[width=.99\linewidth]{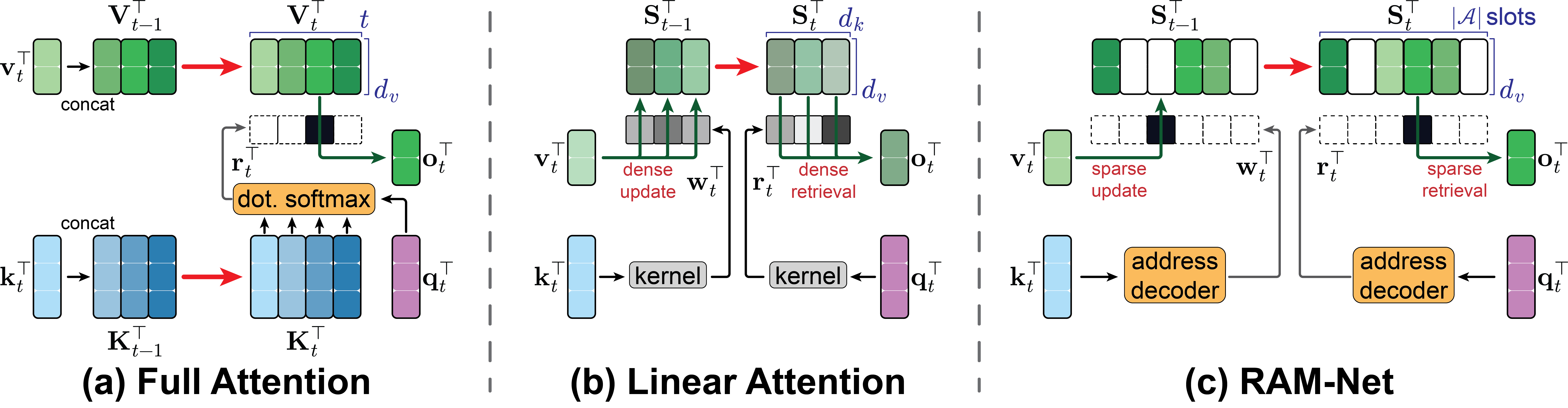}
    \vspace{-0.cm}
    \caption{Comparison of attention architectures. (a) Full attention maintains a KV cache and retrieves through competitive softmax over all cached keys, supporting high retrieval fidelity at the cost of memory that grows with sequence length $t$.
    (b) Linear attention compresses history into a fixed shared recurrent state and retrieves it through dense linear aggregation, thereby causing inter-token interference at retrieval.
    (c) RAM-Net keeps a fixed slot array and uses sparse address-based access: an Address Decoder maps $\mathbf{k}_t$ and $\mathbf{q}_t$ into sparse write and read addresses $\mathbf{w}_t$ and $\mathbf{r}_t$, so only selected slots are updated or retrieved.}
    \label{fig:model_cmp}
\end{figure}

Full attention~\citep{vaswani2017attention} retains the entire history of keys and values (see Fig.~\ref{fig:model_cmp}a).
At step $t$, its state $\mathbf{S}_t = (\mathbf{K}_t, \mathbf{V}_t)$ stores all past keys and values, with $\mathbf{K}_t \in \mathbb{R}^{t \times d_k}$ and $\mathbf{V}_t \in \mathbb{R}^{t \times d_v}$, giving a state size of $t \cdot (d_k + d_v)$.
The write operator simply appends the current key and value, and the read operator scores the query against every cached key with a softmax-normalized similarity, then aggregates the corresponding values:
\begin{equation} \label{eq:full_sdpa}
\begin{aligned}
\mathbf{K}_t &= \text{Concat}(\mathbf{K}_{t-1}, \mathbf{k}_t),
\quad
\mathbf{V}_t = \text{Concat}(\mathbf{V}_{t-1}, \mathbf{v}_t), \\
\mathbf{r}_t &= \text{Softmax}(\mathbf{q}_t \mathbf{K}_t^\top),
\quad~~~
\mathbf{o}_t = \mathbf{r}_t \mathbf{V}_t.
\end{aligned}
\end{equation}
Along the two aspects above, full attention adopts (i) a \textbf{per-token-entry} state granularity, where each token's $(\mathbf{k}_s, \mathbf{v}_s)$ occupies a dedicated entry in $\mathbf{S}_t$, and (ii) a \textbf{softmax-based dense} access pattern, where each query is scored against every cached key via softmax and aggregates the corresponding values.
Full attention achieves high retrieval fidelity through the joint effect of the two choices: per-token entries keep each token's information separable, and softmax concentrates on the most similar entries.
However, the state grows linearly with sequence length and incurs substantial memory and compute cost in long-context inference.

\subsection{Linear Attention}
\label{sec:linear_attn}
Linear attention compresses the state to a fixed size $\mathbf{S}_t \in \mathbb{R}^{d_k \times d_v}$, as shown in Fig.~\ref{fig:model_cmp}(b), giving a state size of $d_k \cdot d_v$.
Vanilla linear attention~\citep{katharopoulos2020transformers} achieves this by replacing the $\exp(\mathbf{q}_t \mathbf{k}_s^\top)$ in softmax with the inner product $\phi(\mathbf{q}_t) \phi(\mathbf{k}_s)^\top$ under a kernel feature map $\phi(\cdot)$.
This approximation factorizes the softmax dependency between query and key, and allows the history to be accumulated into a single matrix.
Concretely, the write applies an outer product $\mathbf{w}_t^\top \mathbf{v}_t$ with $\mathbf{w}_t = \phi(\mathbf{k}_t)$, and the read aggregates $\mathbf{S}_t$ via $\mathbf{r}_t = \phi(\mathbf{q}_t)$:
\begin{equation}
\mathbf{S}_t = \mathbf{S}_{t-1} + \mathbf{w}_t^\top \mathbf{v}_t,
\quad
\mathbf{o}_t \propto \mathbf{r}_t \mathbf{S}_t = \sum\nolimits_{s \leq t} \phi(\mathbf{q}_t) \phi(\mathbf{k}_s)^\top \mathbf{v}_s.
\end{equation}
Modern variants generalize this update to $\mathbf{S}_t = \mathbf{\Lambda}_t \mathbf{S}_{t-1} + \mathbf{w}_t^\top \mathbf{v}_t$, where the transition matrix $\mathbf{\Lambda}_t$ introduces data-dependent decay, gating, or delta-rule corrections~\citep{dao2024transformers, qin2024hgrn2gatedlinearrnns, yang2024gla, yang2024gated}, refining how past contributions are weighted across time while still accumulating them into a single shared matrix.

Examined along the two design aspects above, linear attention adopts (i) an \textbf{unpartitioned matrix} state granularity, where the outer-product update $\mathbf{w}_t^\top \mathbf{v}_t$ writes every token's value into the same $d_k \times d_v$ matrix with no per-token entries, and (ii) an \textbf{additive dense} access pattern, where each step reads from the entire matrix via $\mathbf{r}_t \mathbf{S}_t$, which expands into an additive aggregation $\sum_s \phi(\mathbf{q}_t) \phi(\mathbf{k}_s)^\top \mathbf{v}_s$ over all past values.
Crucially, retrieval interference emerges from the joint effect of the two choices: the unpartitioned matrix superposes contributions from different tokens into the same storage, and the additive read sums all of them into the output with limited ability to suppress the irrelevant ones~\citep{schlag2021linear, sun2024learning}.

Overall, full attention and linear attention occupy two contrasting positions in this two-aspect design space.
Full attention stores tokens as separate entries and weights them through a softmax, which in practice concentrates on a few high-similarity entries~\citep{child2019generating}.
As a result, it achieves high retrieval fidelity, but its state grows with sequence length.
In contrast, linear attention bounds the state by superposing all tokens into a single matrix and reading it through additive aggregation.
This design reduces cost but loses both the separability and the selective access of full attention.
The contrast points to a third paradigm, which keeps the state bounded as in linear attention yet divides it into separable units and accesses only a few of them at each step.
In principle, such a paradigm can retain the efficiency of linear attention while approaching the retrieval fidelity of full attention.
The next section develops RAM-Net within this paradigm.

\section{RAM-Net}
\label{sec:ramnet}
\subsection{Overview: Sparse Addressing on a Slot Array} \label{sec:overview}
The core idea of RAM-Net is twofold: it organizes the fixed-size state into many independent units; at each step, the current key and query each select a sparse subset of these units to write into and read from, respectively.

Along the two aspects in \S\ref{sec:background}, RAM-Net makes the following choices.
(i) For state granularity, RAM-Net organizes the state as an \textbf{addressable slot array}.
The array contains $|\mathcal{A}|$ slots, whose integer addresses form the finite address space $\mathcal{A}=\{0,\ldots,|\mathcal{A}|-1\}$.
At step $t$, the slot at address $a$ stores a content vector $\mathbf{s}_t[a]\in\mathbb{R}^{1\times d_v}$ that aggregates the values written to it so far (\S\ref{sec:gsu}), and stacking all slots gives the state $\mathbf{S}_t\in\mathbb{R}^{|\mathcal{A}|\times d_v}$, as illustrated in Fig.~\ref{fig:model_cmp}(c).
(ii) For the access pattern, RAM-Net adopts \textbf{softmax-based sparse} access.
At each step, the key $\mathbf{k}_t$ produces a write weight vector $\mathbf{w}_t$ over $\mathcal{A}$, and the query $\mathbf{q}_t$ produces a read weight vector $\mathbf{r}_t$.
Each vector has only $K\ll|\mathcal{A}|$ nonzero entries.
The positions of these entries select the slots to write $\mathbf{v}_t$ into or read from, and their magnitudes $w_t[a]$ and $r_t[a]$ specify the write and read strengths at slot $a$.
These weights come from a softmax-based mapping over $\mathcal{A}$ followed by Top-$K$ truncation, as detailed in \S\ref{sec:addr_decode}.

\begin{figure*}
    \centering
    \includegraphics[width=0.99\linewidth]{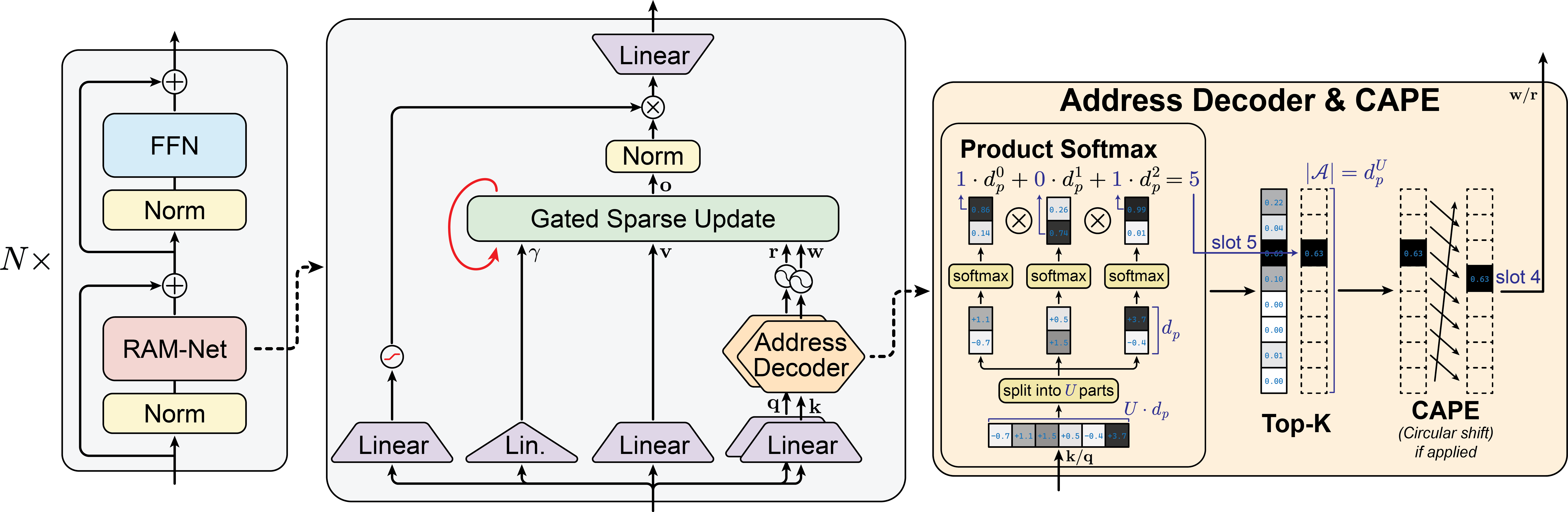}
\caption{Overview of RAM-Net.
The Address Decoder maps the key $\mathbf{k}_t$ and query $\mathbf{q}_t$ to sparse weights over the slots through Product Softmax and Top-$K$ truncation.
Cyclic Address Positional Embedding (CAPE) then optionally shifts their slot indices by token position, producing the write and read weights $\mathbf{w}_t$ and $\mathbf{r}_t$.
Next, Gated Sparse Update writes $\mathbf{v}_t$ into the selected slots, and the read output $\mathbf{o}_t$ is normalized, gated, and projected.
For visual clarity, the Address Decoder panel illustrates $K=1$ and $|\mathcal{A}|=8$ slots, with CAPE shifting the selected slot 5 to slot 4.
}
    \label{fig:arch}
\end{figure*}

\subsection{Address Decoder}
\label{sec:addr_decode}
\begin{figure*}
    \centering
    \includegraphics[width=0.99\linewidth]{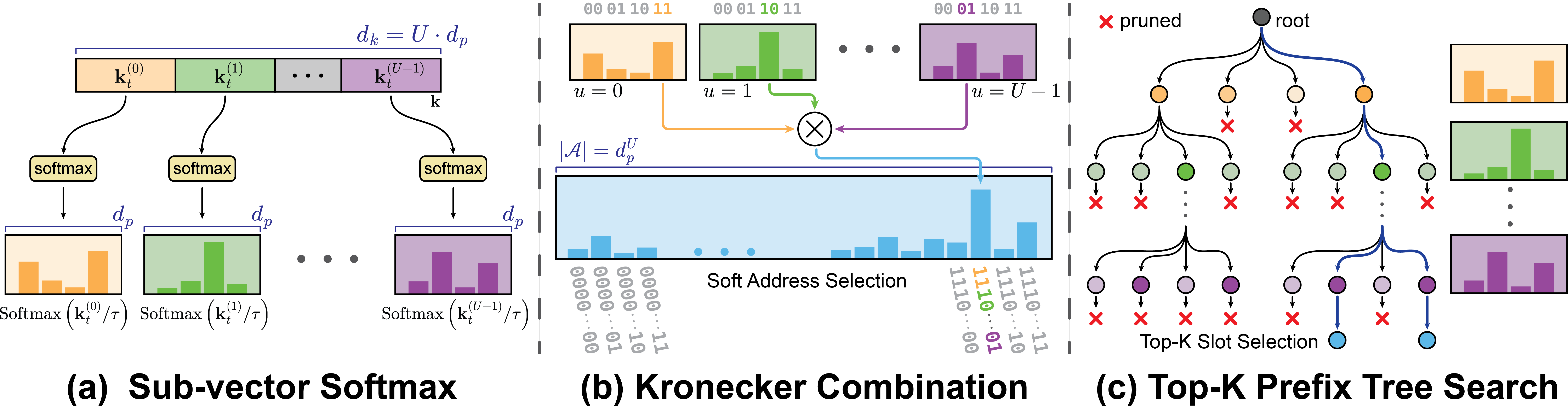}
    \caption{Address Decoder.
(a) The key $\mathbf{k}_t$ or query $\mathbf{q}_t$ is split into $U$ sub-vectors of size $d_p$, and each passes through its own softmax to give one digit distribution.
(b) The Kronecker product of the digit distributions gives the address distribution over $|\mathcal{A}|=d_p^U$ slots.
(c) A Top-$K$ tree search keeps the $K$ most probable partial addresses at each level, returning the exact Top-$K$ slots without materializing the $|\mathcal{A}|$-dimensional distribution.}
    \label{fig:product_softmax}
\end{figure*}

This section presents the Address Decoder, which maps each key or query to sparse weights over the slots to write to or read from at the current step (Fig.~\ref{fig:arch}).
To obtain selective access as in full attention, a direct realization would project keys and queries to $|\mathcal{A}|$ dimensions, apply a softmax over all slots, and retain only the top-$K$ weights.
However, this construction faces two obstacles as $|\mathcal{A}|$ grows.
First, the projections and the $|\mathcal{A}|$-dimensional softmax add parameters and computation that grow linearly with $|\mathcal{A}|$.
Second, after Top-$K$ truncation, only the $K$ selected entries out of $|\mathcal{A}|$ receive a direct learning signal at each step, so the address selection becomes hard to optimize as $|\mathcal{A}|$ grows (Appendix~\ref{sec:abla_prod}).
To resolve both obstacles, the Address Decoder combines a \textit{Product Softmax} construction with a Top-$K$ truncation step, as illustrated in Fig.~\ref{fig:product_softmax}.

\paragraph{Product Softmax.}
Product Softmax $\rho_U(\cdot)$ builds the distribution over all $|\mathcal{A}|$ slots from $U$ small softmax distributions.
Specifically, as illustrated in Fig.~\ref{fig:product_softmax}(a), the key $\mathbf{k}_t$ (and similarly $\mathbf{q}_t$) is split into $U$ sub-vectors $\mathbf{k}_t=[\mathbf{k}_t^{(0)},\ldots,\mathbf{k}_t^{(U-1)}]$, each of dimension $d_p=d_k/U$.
We set $|\mathcal{A}|=d_p^U$, so that each address $a$ can be written as \textbf{$U$ digits in radix $d_p$}, $a=\sum_{u=0}^{U-1}a^{(u)}d_p^{u}$ with $a^{(u)}\in\{0,\ldots,d_p-1\}$.
Each sub-vector $\mathbf{k}_t^{(u)}$ scores the $d_p$ choices of digit $a^{(u)}$, and a softmax turns these scores into a digit distribution $\mathbf{p}_t^{(u)}=\operatorname{Softmax}(\mathbf{k}_t^{(u)}/\tau)$, where $\tau>0$ is a temperature.
As shown in Fig.~\ref{fig:product_softmax}(b), the probability $\rho_U(\mathbf{k}_t)[a]$ of each address $a$ is the product of its digit probabilities, so the full distribution $\rho_U(\mathbf{k}_t)$ is the Kronecker product $\otimes$ of the digit distributions:
\begin{equation}
    \rho_U(\mathbf{k}_t)[a]
    = \prod_{u=0}^{U-1} p_t^{(u)}[a^{(u)}],
    \qquad
    \rho_U(\mathbf{k}_t)
    = \mathbf{p}_t^{(U-1)} \otimes \cdots \otimes \mathbf{p}_t^{(0)}.
\end{equation}

\paragraph{Top-$K$ Truncation.}
To obtain $K$-sparse weights, we apply a Top-$K$ truncation operator $\mathcal{T}_K(\cdot)$, which retains the $K$ largest probabilities and sets the rest to zero.
The complete address decoding is therefore defined as:
\begin{equation}
    \mathcal{D}_{K,U}(\mathbf{k}_t) = \mathcal{T}_K \left( \rho_U(\mathbf{k}_t) \right).
\end{equation}

In practice, we fuse Product Softmax and Top-$K$ truncation, so that the full $|\mathcal{A}|$-dimensional distribution is never materialized.
Expanding the Kronecker product factor by factor yields an implicit prefix tree of depth $U$, with complete addresses at the leaves and each prefix weighted by the product of digit probabilities along its path from the root (Fig.~\ref{fig:product_softmax}(c)).
At each level, we extend each retained prefix with all $d_p$ possible next digits and keep the $K$ most probable extensions.
This pruning is exact because suffix probabilities are independent of the prefix: for any completion of a discarded prefix, appending the same suffix to the $K$ retained prefixes yields $K$ addresses that are at least as probable.
As each level examines only $Kd_p$ candidates, the cost grows linearly in $U=\log_{d_p}|\mathcal{A}|$ rather than in $|\mathcal{A}|$ (see \S\ref{sec:impl} for details).

The Address Decoder thus resolves both obstacles of a direct $|\mathcal{A}|$-dimensional softmax.
For the cost obstacle, with a fixed radix $d_p$, the slot count $|\mathcal{A}|=d_p^U$ grows exponentially in the number of digits $U$, while the key and query dimension $d_k=Ud_p$ grows only linearly, and the Top-$K$ prefix tree search avoids any $|\mathcal{A}|$-dimensional computation.
For the optimization obstacle, each digit is shared by many slots, so the learning signal from a selected slot also reaches every slot that shares its digits.
Consistently, our ablation shows that a larger $U$ improves associative recall accuracy, with larger gains when $|\mathcal{A}|$ is large (Appendix~\ref{sec:ablation}).


\subsection{Cyclic Address Positional Embedding}
\label{sec:cape}
The Address Decoder computes addresses from content alone, so identical content maps to the same slots regardless of its position.
To inject positional information, we propose \textit{Cyclic Address Positional Embedding (CAPE)}, which cyclically shifts the decoded addresses according to the token position, as illustrated in the CAPE block of Fig.~\ref{fig:arch}.
Specifically, CAPE moves each selected address $a$ to $\pi_t(a)=(a-t)\bmod|\mathcal{A}|$ while keeping its weight, which changes the slots being accessed but not their contents.
The final write and read weights are thus given by
\begin{equation} \label{equ:wr}
    w_t[\pi_t(a)] = \mathcal{D}_{K,U}(\mathbf{k}_t)[a], \qquad
    r_t[\pi_t(a)] = \mathcal{D}_{K,U}(\mathbf{q}_t)[a], \qquad a\in\mathcal{A}.
\end{equation}
Since writes at step $s$ are shifted by $s$ and reads at step $t$ by $t$, the overlap between $\mathbf{r}_t$ and an earlier $\mathbf{w}_s$ depends jointly on their content and the relative distance $t-s$.
This is analogous to RoPE~\citep{su2024roformer}, which obtains relative positions by rotating queries and keys, whereas CAPE does so by shifting slot addresses.
Since position-aware and content-only addressing capture different patterns, RAM-Net applies CAPE to some heads and No Positional Embedding (NoPE) to the others (see Appendix \ref{sec:training_details} for the configuration and Appendix~\ref{appendix:access_pattern} for access visualizations).

\subsection{Gated Sparse Update}
\label{sec:gsu}
A state update must support both writing new information and forgetting outdated content.
In gated linear-attention models, forgetting is implemented through a gate applied to the entire state at each step~\citep{dao2024transformers, qin2024hgrn2gatedlinearrnns, yang2024gla, yang2024gated}.
Applying such a gate to the slot array would access all $|\mathcal{A}|$ slots per step.
This conflicts with sparse addressing, which modifies only the slots selected by $\mathbf{w}_t$ and leaves all other slots exactly unchanged (\S\ref{sec:overview}).
We therefore propose \textit{Gated Sparse Update} (GSU), which applies a forgetting gate only to the addressed slots, so that writing and forgetting both remain sparse.

GSU uses a \emph{per-slot test-time training} (TTT) update~\citep{sun2024learning}, taking a gradient step on the squared-error objective $\ell_t(\mathbf{s})=\tfrac12\|\mathbf{s}-\mathbf{v}_t\|_2^2$ with an adaptive learning rate.
Each slot $a$ maintains a content vector $\mathbf{s}_t[a]$ and a scalar mass $m_t[a]$ representing its inertia against new writes.
At each step, GSU first scales the previous mass by a retention factor $\lambda_t[a]$ and adds the current write weight $w_t[a]$ from Eq.~\eqref{equ:wr}:
\begin{equation}
\begin{aligned}
    \lambda_t[a] &= \sigma\!\left(\mathrm{logit}(1-w_t[a])-\gamma_t\right),\\
    m_t[a] &= \lambda_t[a]\,m_{t-1}[a]+w_t[a],
\end{aligned}
\label{eq:retention}
\end{equation}
where $\sigma$ is the sigmoid function and $\mathrm{logit}(x)=\log\frac{x}{1-x}$.
The retention factor uses $1-w_t[a]$ as its baseline, so a slot discards more of its mass when written more strongly and keeps all of it when not written.
A forgetting modulator $\gamma_t$ adjusts this baseline in logit space. It is computed from $\mathbf{x}_t$ by a learned per-head projection, giving one scalar per token and head shared across slots (see Fig.~\ref{fig:gamma} in Appendix~\ref{sec:training_details}).
Then, GSU uses the updated mass to compute the learning rate $\eta_t[a]$ and update the slot content:
\begin{equation}
\begin{aligned}
    \eta_t[a] &= \frac{w_t[a]}{m_t[a]+\epsilon},\\
    \mathbf{s}_t[a] &= \mathbf{s}_{t-1}[a]+\eta_t[a]\bigl(\mathbf{v}_t-\mathbf{s}_{t-1}[a]\bigr),
\end{aligned}
\label{eq:gsu}
\end{equation}
with $m_0[a]=1/|\mathcal{A}|$, $\mathbf{s}_0[a]=\mathbf{0}$, and a small constant $\epsilon>0$ for numerical stability.

These updates give GSU three properties.
(i) \textbf{Exact sparsity.}
For an unaddressed slot, $w_t[a]=0$ gives $\lambda_t[a]=1$ by continuity and $\eta_t[a]=0$.
Therefore, both its mass and content remain exactly unchanged.
Unaddressed slots thus need not be accessed at all, and each step accesses only the slots selected by $\mathbf{w}_t$, as required by sparse addressing.
(ii) \textbf{Bounded content.}
Since $\lambda_t[a]\,m_{t-1}[a]\geq0$, the mass satisfies $m_t[a]\geq w_t[a]\geq0$, and with $\epsilon>0$ the learning rate satisfies $\eta_t[a]\in[0,1)$.
Each content update is therefore a convex combination of the previous content and $\mathbf{v}_t$.
Every slot thus stays within the convex hull of its initial content and the values written to it, so its norm never exceeds the largest norm among these values, regardless of how often it is updated.
(iii) \textbf{Adaptive learning rate.}
The learning rate $\eta_t[a]$ is adapted at two levels.
First, each slot's learning rate depends on its current write weight $w_t[a]$ relative to its historical mass $m_{t-1}[a]$.
A stronger write increases the learning rate, whereas a larger historical mass provides greater inertia and suppresses the update.
Second, the input-dependent modulator $\gamma_t$ controls how much of this historical mass is retained through the retention factor $\lambda_t[a]$.
This allows the model to balance preserving history against adopting new content for each token and head.
As $\gamma_t\to-\infty$, $\lambda_t[a]\to1$ and the slot becomes a write-weighted running average of the values written to it.
As $\gamma_t\to+\infty$, $\lambda_t[a]\to0$ and $\eta_t[a]\approx1$, so each write nearly overwrites the slot.

\paragraph{Read-out.}
The read weights retrieve $\mathbf{o}_t=\mathbf{r}_t\mathbf{S}_t=\sum_{a\in\mathcal{A}}r_t[a]\mathbf{s}_t[a]$ from the selected slots.
Finally, $\mathbf{o}_t$ passes through a head-wise RMSNorm~\citep{rms}, a sigmoid output gate, and the output projection, giving $\mathbf{y}_t=\big(\sigma(\mathbf{x}_tW_g)\odot\mathrm{RMSNorm}(\mathbf{o}_t)\big)W_o$, as shown in Fig.~\ref{fig:arch}.

\section{Implementation}
\label{sec:impl}

\paragraph{Efficient Address Decoding.}
For each token and head, the write and read addresses are decoded separately by the Top-$K$ tree search in \S\ref{sec:addr_decode}.
We first sort the $d_p$ logits and compute their softmax for each of the $U$ digit distributions entirely in registers, at a total cost of $\mathcal{O}(Ud_p\log d_p)$.
The search then proceeds level by level, using a wavefront scan to combine the $K$ retained prefixes with the $d_p$ choices for the next digit and select the $K$ most probable extensions, at a cost of $\mathcal{O}(Kd_p)$ per level.
The total per-step cost is therefore $\mathcal{O}(UKd_p+Ud_p\log d_p)$, far below the $\mathcal{O}(|\mathcal{A}|)$ cost of an exhaustive search.
Since an address probability is a product of $U$ digit probabilities, all computations are carried out in log space, which turns the product into a sum and avoids numerical underflow.

\paragraph{Slot-Parallel Execution.}
Although each step accesses only a few slots, the selected slots are scattered across the state and differ from token to token, which makes the accesses irregular on GPUs.
During training and prefill, we therefore organize the computation by slot rather than by token.
First, the write and read addresses of all tokens are decoded in parallel.
Next, all read and write events are sorted by slot, so that each slot receives a time-ordered list of the events that access it, and slots without events are skipped.
Finally, different slots and the $d_v$ channels within each slot are processed in parallel.
Within each slot, the events are applied in time order as required by the recurrence, and each read result is accumulated back to its token.
The total work thus scales with the number of events, which is proportional to the sequence length times $K$, rather than with the full state size.
During autoregressive decoding, each new token simply updates and reads its own $K$ selected slots.

\section{Experiments} \label{sec:exp}
\paragraph{Baselines and Experimental Setup.}
We compare RAM-Net against Transformer++ \citep{transformer_llama}, Mamba2 \citep{dao2024transformers}, GLA \citep{yang2024gla}, GSA~\citep{zhang2024gsa}, DeltaNet \citep{schlag2021linear}, MoM \citep{du2025mom}, and Gated DeltaNet \citep{yang2024gated}, with all baselines using the implementations provided by the Flash-Linear-Attention library \citep{yang2024fla}. 
All models are trained using the Flame framework \citep{yang2025flame} on the FineWeb-Edu dataset \citep{lozhkov2024fineweb-edu} with the Llama2 tokenizer \citep{touvron2023llama} and a context window of 4,096 tokens. 
We train models at two scales: 340M parameters on 10B tokens, and 1.3B parameters on 100B tokens. 
For both scales, we use AdamW optimizer \citep{loshchilov2019decoupledweightdecayregularization} with weight decay of $0.1$, gradient clipping of $1.0$, cosine decay schedule, and a global batch size of 0.5M tokens.
The peak learning rate and warmup tokens are set to $1.0 \times 10^{-3}$ and 500M for the 340M scale, and $4.0 \times 10^{-4}$ and 1B for the 1.3B scale, 
respectively (see training details in Appendix \ref{sec:training_details}).
The ablation studies are reported in Appendix \ref{sec:ablation}.

\begin{table*}[t]
\centering
\caption{Zero-shot performance on language modeling and commonsense reasoning. }
\vspace{-0.15cm}
\label{tab:main_results}
\resizebox{\textwidth}{!}{
\begin{tabular}{l|cc|ccccccccccc|c}
\toprule
\textbf{Model} &
\shortstack{\textbf{Wiki.}\\\footnotesize $\text{ppl}$ $\downarrow$} &
\shortstack{\textbf{LMB.}\\\footnotesize $\text{acc}$ $\uparrow$} &
\shortstack{\textbf{MMLU}\\\footnotesize $\text{acc}$ $\uparrow$} &
\shortstack{\textbf{ARC-e}\\\footnotesize $\text{acc}$ $\uparrow$} &
\shortstack{\textbf{ARC-c}\\\footnotesize $\text{acc}$ $\uparrow$} &
\shortstack{\textbf{OBQA}\\\footnotesize $\text{acc}_\text{n}$ $\uparrow$} &
\shortstack{\textbf{SciQ}\\\footnotesize $\text{acc}$ $\uparrow$} &
\shortstack{\textbf{BoolQ}\\\footnotesize $\text{acc}$ $\uparrow$} &
\shortstack{\textbf{COPA}\\\footnotesize $\text{acc}$ $\uparrow$} &
\shortstack{\textbf{PIQA}\\\footnotesize $\text{acc}$ $\uparrow$} &
\shortstack{\textbf{Hella.}\\\footnotesize $\text{acc}$ $\uparrow$} &
\shortstack{\textbf{Wino.}\\\footnotesize $\text{acc}$ $\uparrow$} &
\shortstack{\textbf{Avg.}\\\footnotesize $\text{acc}$ $\uparrow$} &
\shortstack{\textbf{Act. State}\textsuperscript{1}\\\footnotesize per step $\downarrow$} \\
\midrule
\multicolumn{15}{l}{\textit{340M}} \\
Transformer++ & 26.81 & \ul{33.1} & 23.1 & 57.7 & 24.7 & 32.0 & 82.9 & 55.9 & 66.0 & 65.8 & 33.3 & 52.2 & 49.4 & 100.7M \\
Gated DeltaNet & 26.85 & 32.4 & 23.4 & 57.9 & 25.0 & \ul{35.0} & \ul{83.3} & 58.9 & 67.0 & 66.8 & \tb{33.9} & 50.2 & 50.1 & 4.4M \\
DeltaNet & 27.61 & 32.0 & 22.9 & 58.1 & 24.9 & 31.6 & 80.0 & \ul{61.2} & \tb{71.0} & 65.5 & 33.2 & 50.0 & 49.8 & 3.4M \\
Mamba2 & \ul{26.23} & \tb{33.9} & 23.1 & 59.2 & 24.2 & 33.8 & \tb{83.4} & 59.6 & \tb{71.0} & \tb{67.7} & \tb{33.9} & 50.4 & \ul{50.6} & 13.0M \\
GLA & 30.03 & 30.4 & 23.2 & 56.9 & 25.0 & 32.4 & 81.6 & 56.8 & 66.0 & 66.4 & 32.7 & 50.7 & 49.2 & 3.1M \\
GSA & 29.17 & 32.7 & 23.0 & \ul{59.3} & 25.8 & \tb{35.8} & 83.0 & 57.4 & 66.0 & 66.4 & 33.3 & 49.8 & 50.0 & 3.1M \\
MoM & 28.27 & 30.2 & 22.9 & 56.8 & \ul{25.9} & 31.8 & 82.5 & 56.3 & 66.0 & 65.4 & 32.8 & 51.0 & 49.1 & 5.2M \\
\rowcolor{ramrow}
RAM-Net & & & & & & & & & & & & & & \\
\rowcolor{ramrow}
\quad Top-8 & 26.89 & 28.7 & \ul{23.6} & 59.1 & 24.7 & 34.4 & 82.5 & 57.4 & 67.0 & 66.9 & 33.4 & \tb{54.1} & 50.3 & \tb{0.4M} \\
\rowcolor{ramrow}
\quad Top-16 & \ul{26.23} & 29.7 & 23.5 & \tb{59.4} & \tb{26.1} & 33.0 & 82.8 & 58.6 & 69.0 & \ul{67.2} & \ul{33.8} & 51.4 & 50.5 & \ul{0.8M} \\
\rowcolor{ramrow}
\quad Top-32 & \tb{25.87} & 30.9 & \tb{23.8} & 58.9 & \tb{26.1} & 32.2 & 82.8 & \tb{61.6} & \ul{70.0} & \ul{67.2} & 33.7 & \ul{52.7} & \tb{50.9} & 1.7M \\
\midrule
\multicolumn{15}{l}{\textit{1.3B}} \\
Transformer++ & 17.08 & 46.9 & 23.0 & 70.2 & 34.0 & 39.2 & 88.7 & 59.1 & 73.0 & 71.3 & 42.8 & \ul{58.7} & 56.0 & 201.3M \\
Gated DeltaNet & \tb{15.85} & \tb{50.1} & \tb{25.5} & \ul{72.4} & \tb{38.7} & 42.0 & \tb{91.9} & \tb{63.6} & 76.0 & \ul{73.5} & \ul{45.2} & 58.2 & \tb{58.7} & 17.0M \\
Mamba2 & 15.96 & \ul{49.9} & 24.5 & 71.9 & 36.3 & \ul{42.6} & 88.9 & 55.1 & \tb{81.0} & \ul{73.5} & \tb{45.6} & 58.2 & 57.8 & 26.0M \\
\rowcolor{ramrow}
RAM-Net & & & & & & & & & & & & & & \\
\rowcolor{ramrow}
\quad Top-8 & 16.49 & 43.0 & 23.6 & 72.0 & 36.3 & 41.8 & 88.2 & \ul{60.3} & 76.0 & 73.0 & 44.6 & 57.9 & 57.4 & \tb{0.8M} \\
\rowcolor{ramrow}
\quad Top-16 & \ul{15.90} & 47.4 & 23.6 & 71.8 & \ul{37.6} & 40.8 & 89.6 & 58.3 & \ul{79.0} & 72.9 & 44.9 & \tb{59.1} & 57.8 & \ul{1.7M} \\
\rowcolor{ramrow}
\quad Top-32 & 16.02 & 46.7 & \ul{24.8} & \tb{72.8} & 36.6 & \tb{43.6} & \ul{90.6} & 57.2 & \ul{79.0} & \tb{73.7} & 44.9 & 57.8 & \ul{58.1} & 3.3M \\
\bottomrule
\end{tabular}
}
\vspace{1mm}
\parbox{\textwidth}{%
    \scriptsize
    \raggedright
\textsuperscript{1} denotes the state size accessed per step; Transformer++ is measured at context length 2048. This equals the total state size for all models except MoM and RAM-Net, whose total state sizes are 8.6M for MoM at 340M and 26.6M/53.2M for RAM-Net at 340M/1.3B.
}
\vspace{-0.8cm}
\end{table*}

\paragraph{Language Modeling and Commonsense Reasoning.}
We evaluate zero-shot language modeling and commonsense reasoning with the Language Model Evaluation Harness~\citep{eval-harness} on WikiText-103~\citep{merity2016pointersentinelmixturemodels}, 
LAMBADA (LMB)~\citep{paperno2016lambadadatasetwordprediction}, MMLU~\citep{hendrycks2021measuringmassivemultitasklanguage}, ARC-Challenge/Easy~\citep{clark2018thinksolvedquestionanswering}, OpenBookQA~\citep{mihaylov2018can}, 
SciQ~\citep{welbl2017crowdsourcingmultiplechoicescience}, BoolQ~\citep{clark2019boolq}, COPA~\citep{roemmele2011choice}, PIQA~\citep{bisk2020piqa}, HellaSwag~\citep{zellers2019hellaswagmachinereallyfinish}, and WinoGrande~\citep{sakaguchi2019winograndeadversarialwinogradschema}. 
The results are summarized in Table~\ref{tab:main_results}.
First, RAM-Net achieves the lowest WikiText perplexity at 340M and performs on par with the strongest baseline Gated DeltaNet at 1.3B.
Second, its commonsense reasoning remains competitive. RAM-Net Top-32 obtains the highest average accuracy at 340M and the second highest at 1.3B.
Third, increasing Top-$K$ consistently improves average accuracy at both scales, suggesting that accessing more slots per step strengthens commonsense reasoning.
Notably, RAM-Net achieves these results while accessing far fewer state elements per step than every baseline.

\begin{wrapfigure}{r}{0.45\textwidth}
    \centering
    \includegraphics[width=1\linewidth]{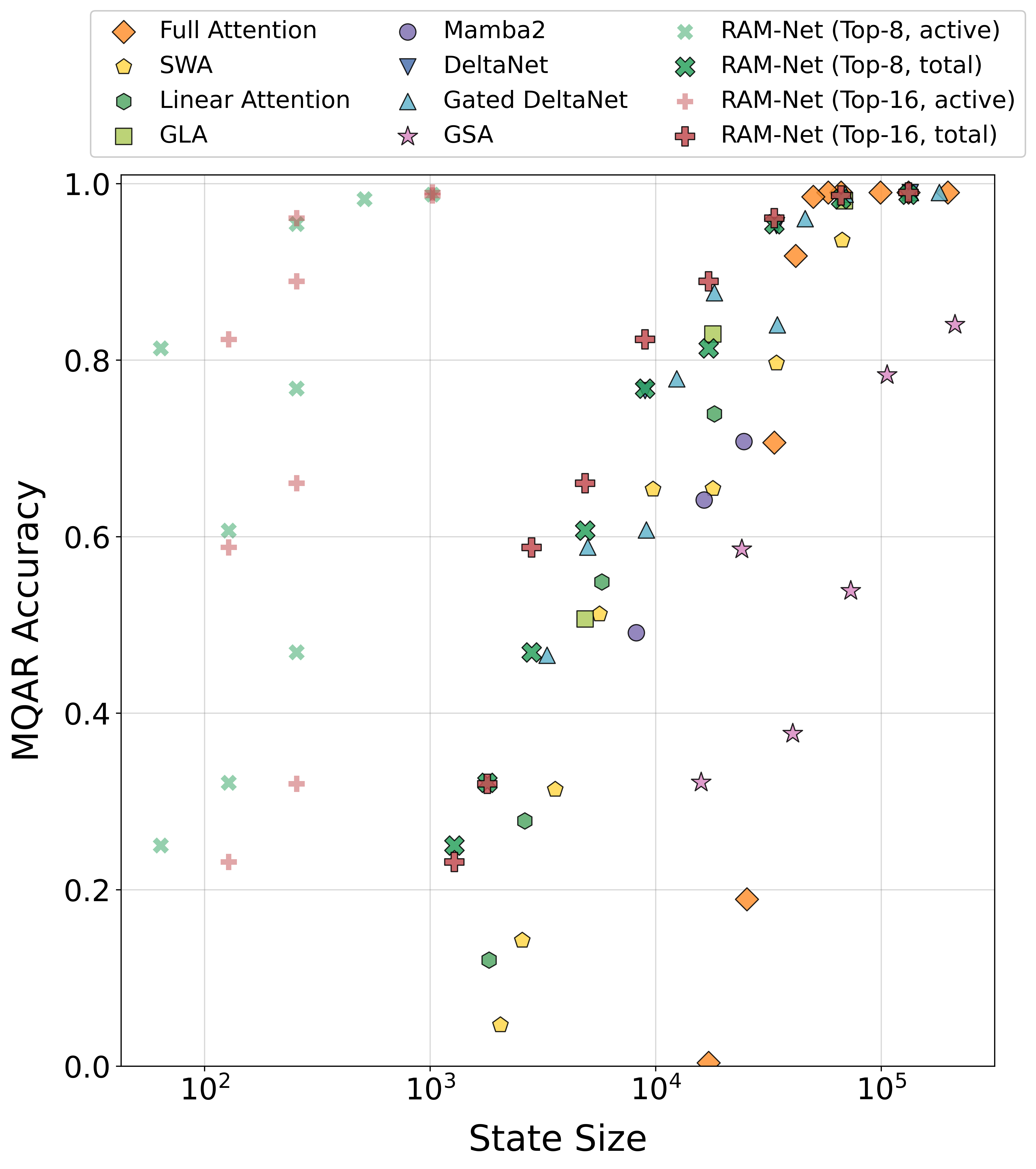}
    \caption{MQAR accuracy versus state size. Active state in RAM-Net denotes the effective state accessed per step.}
    \label{fig:mqar_result}
    \vspace{-0.5cm}
\end{wrapfigure}

\paragraph{Synthetic Associative Recall.}
We evaluate RAM-Net on multi-query associative recall (MQAR) \citep{arora2024zoology} using Zoology, a synthetic benchmark that tests whether a model can store key-value pairs and retrieve the correct value from a query key under controlled conditions. 
To focus on architectural recall capacity, we train all models from scratch for this task.
Beyond the baselines above, we additionally include linear attention \citep{katharopoulos2020transformers} and sliding window attention (SWA) \citep{beltagy2020longformer}.
We follow the Zoology evaluation protocol and report average accuracy over the standard MQAR settings.
As shown in Fig.~\ref{fig:mqar_result}, RAM-Net (Top-8/16) lies on the upper envelope against total state size, reaching high accuracy at smaller total states than most linear recurrent baselines.
Moreover, RAM-Net reaches this accuracy while accessing only a small fraction of its state per step, about $10\times$ less than baselines of comparable accuracy.
These results suggest that sparse addressing makes more effective use of the total state, and that restricting each step to a small part of it does not compromise recall.

\begin{table*}[t]
\centering
\caption{Zero-shot performance comparison on S-NIAH benchmark. All models are trained with a 4K context window.}
\label{tab:niah}
\vspace{-0.5mm}
\scriptsize
\setlength{\tabcolsep}{3.5pt}
\renewcommand{\arraystretch}{1.0}
\begin{tabular}{l|ccccc|ccccc|ccccc|c}
\toprule
& \multicolumn{5}{c|}{{S-NIAH-1}} & \multicolumn{5}{c|}{{S-NIAH-2}} & \multicolumn{5}{c|}{{S-NIAH-3}} & \\
\cmidrule(lr){2-6} \cmidrule(lr){7-11} \cmidrule(lr){12-16}
{Model} & {1K} & {2K} & {4K} & {8K} & {16K} &
{1K} & {2K} & {4K} & {8K} & {16K} &
{1K} & {2K} & {4K} & {8K} & {16K} & {Avg.} \\
\midrule
\multicolumn{17}{l}{\textit{340M}} \\
Transformer++ & \ul{99.7} & 97.7 & 81.0 & 0.0 & 0.0 & \tb{100.0} & \ul{99.7} & 74.7 & 0.0 & 0.0 & \ul{81.7} & 62.7 & \ul{31.3} & 0.0 & 0.0 & 48.6 \\
Gated DeltaNet & 99.3 & 99.3 & 98.7 & 88.0 & 44.0 & \ul{99.7} & 96.0 & 37.3 & 20.3 & \tb{10.3} & 46.0 & 26.7 & 2.3 & 0.7 & 0.3 & 51.3 \\
DeltaNet & \tb{100.0} & \tb{100.0} & \tb{100.0} & \ul{99.7} & 98.7 & 94.3 & 95.7 & 18.0 & 23.0 & \ul{7.0} & 48.7 & 21.7 & 7.0 & 2.7 & \tb{1.3} & 54.5 \\
Mamba2 & 89.3 & 91.7 & 77.0 & 24.7 & 6.7 & 97.3 & 72.7 & 12.0 & 11.7 & 4.0 & 20.7 & 10.3 & 5.3 & 1.3 & 0.7 & 35.0 \\
GLA & 68.7 & 23.7 & 5.7 & 0.3 & 0.3 & 64.7 & 20.3 & 6.3 & 3.3 & 1.0 & 5.0 & 0.3 & 0.0 & 0.0 & 0.0 & 13.3 \\
GSA & 75.3 & 54.7 & 31.7 & 13.7 & 7.3 & 69.0 & 49.7 & 11.0 & 7.0 & 1.7 & 0.0 & 0.0 & 0.0 & 0.0 & 0.0 & 21.4 \\
MoM & \tb{100.0} & \ul{99.7} & 98.7 & 89.7 & 59.0 & 97.0 & 76.0 & 26.3 & 10.0 & 0.7 & 19.3 & 8.0 & 1.7 & 0.3 & 0.3 & 45.8 \\
\rowcolor{ramrow}
RAM-Net & & & & & & & & & & & & & & & & \\
\rowcolor{ramrow}
\quad Top-8 & \tb{100.0} & \tb{100.0} & \ul{99.7} & \ul{99.7} & \ul{99.7} & 99.3 & 99.3 & 83.7 & 20.7 & 5.3 & 69.3 & \ul{69.3} & 18.3 & 5.7 & \ul{1.0} & 64.7 \\
\rowcolor{ramrow}
\quad Top-16 & \tb{100.0} & \tb{100.0} & \tb{100.0} & \tb{100.0} & \tb{100.0} & \tb{100.0} & \ul{99.7} & \ul{88.0} & \ul{23.3} & 2.7 & 70.7 & 58.0 & 25.7 & \ul{7.3} & \tb{1.3} & \ul{65.1} \\
\rowcolor{ramrow}
\quad Top-32 & \tb{100.0} & \tb{100.0} & \tb{100.0} & \tb{100.0} & \tb{100.0} & \tb{100.0} & \tb{100.0} & \tb{94.0} & \tb{56.3} & 4.0 & \tb{95.3} & \tb{88.0} & \tb{44.7} & \tb{9.7} & \ul{1.0} & \tb{72.9} \\
\midrule
\multicolumn{17}{l}{\textit{1.3B}} \\
Transformer++ & \tb{100.0} & \tb{100.0} & 95.0 & 0.0 & 0.0 & \tb{100.0} & \tb{100.0} & \tb{90.5} & 4.0 & 0.0 & \tb{98.0} & \tb{91.6} & \tb{66.0} & 0.4 & 0.0 & 56.4 \\
Gated DeltaNet & \tb{100.0} & \ul{99.7} & 99.0 & 61.7 & 29.0 & \tb{100.0} & 99.3 & 77.0 & 30.0 & \ul{14.3} & 72.7 & 62.7 & 20.3 & \ul{9.7} & \ul{4.3} & 58.6 \\
Mamba2 & \tb{100.0} & \tb{100.0} & \tb{100.0} & \ul{99.7} & 71.0 & 99.0 & 95.0 & 65.7 & 17.0 & 5.7 & \ul{89.7} & 69.7 & 15.0 & 6.3 & 2.3 & 62.4 \\
\rowcolor{ramrow}
RAM-Net & & & & & & & & & & & & & & & & \\
\rowcolor{ramrow}
\quad Top-8 & \tb{100.0} & \tb{100.0} & \tb{100.0} & \tb{100.0} & \tb{100.0} & \ul{99.7} & \ul{99.7} & 82.7 & \ul{42.3} & 9.3 & 63.0 & 54.0 & 16.0 & 8.3 & 0.3 & 65.0 \\
\rowcolor{ramrow}
\quad Top-16 & \tb{100.0} & \ul{99.7} & \ul{99.7} & \ul{99.7} & \ul{99.7} & \tb{100.0} & 97.7 & \ul{85.7} & \tb{49.0} & \tb{18.7} & 74.7 & 63.3 & \ul{39.0} & \tb{18.3} & \tb{6.3} & \tb{70.1} \\
\rowcolor{ramrow}
\quad Top-32 & \ul{99.3} & \tb{100.0} & \tb{100.0} & \tb{100.0} & \tb{100.0} & \tb{100.0} & 98.0 & 80.7 & 33.3 & 10.7 & 78.0 & \ul{73.3} & 21.0 & 9.0 & 1.7 & \ul{67.0} \\
\bottomrule
\end{tabular}
\vspace{-0.5cm}
\end{table*}

\paragraph{Long-Context Retrieval.}
We evaluate long-context retrieval on the single needle-in-a-haystack (S-NIAH) tasks from RULER~\citep{hsieh2024ruler}{\renewcommand{\thefootnote}{\ddag}\footnote{\label{fn:cloze} All models are pretrained only with next-token prediction, so we evaluate S-NIAH and the recall tasks as cloze-style completion~\citep{hsieh2024ruler, arora2024cloze}.}} with context lengths from 1K to 16K, up to four times the 4K training length (Table~\ref{tab:niah}).
RAM-Net achieves the highest average accuracy at both scales.
On S-NIAH-1, where the needle is hidden among repeated filler sentences, all RAM-Net variants remain near-perfect up to 16K tokens.
In contrast, Transformer++ fails beyond its training length, and most recurrent baselines degrade.
We attribute this robustness partly to sparse addressing, as content-only heads write the repeated filler into the same few slots and thus leave the slots holding the needle largely undisturbed.
Unlike S-NIAH-1, S-NIAH-2/3 use diverse natural text as filler and are therefore harder.
On these tasks, Top-32 performs best up to 8K tokens at 340M; at 1.3B, Transformer++ leads within its training length, whereas Top-16 performs best beyond it, although all models remain weak at 16K.
Finally, Top-16 and Top-32 both achieve higher average accuracy than Top-8 at both scales, suggesting that accessing more slots per step benefits retrieval.


\paragraph{Real-World Recall.}
We further evaluate in-context recall on the recall-intensive tasks~\citep{arora2024cloze}\footref{fn:cloze},  which require extracting information from real documents.
As shown in Table~\ref{tab:jrt}, Transformer++ achieves the highest average at both scales, since it stores every context token in its KV cache and can retrieve from any of them directly.
Among recurrent models, 
RAM-Net Top-32 attains the highest average at both scales, and at 340M it narrows the gap to Transformer++ from 8.2 points for the best baseline to 3.1 points.

\paragraph{Decoding Throughput.}
\begin{wrapfigure}{r}{0.31\textwidth}
    \centering
    \includegraphics[width=1\linewidth]{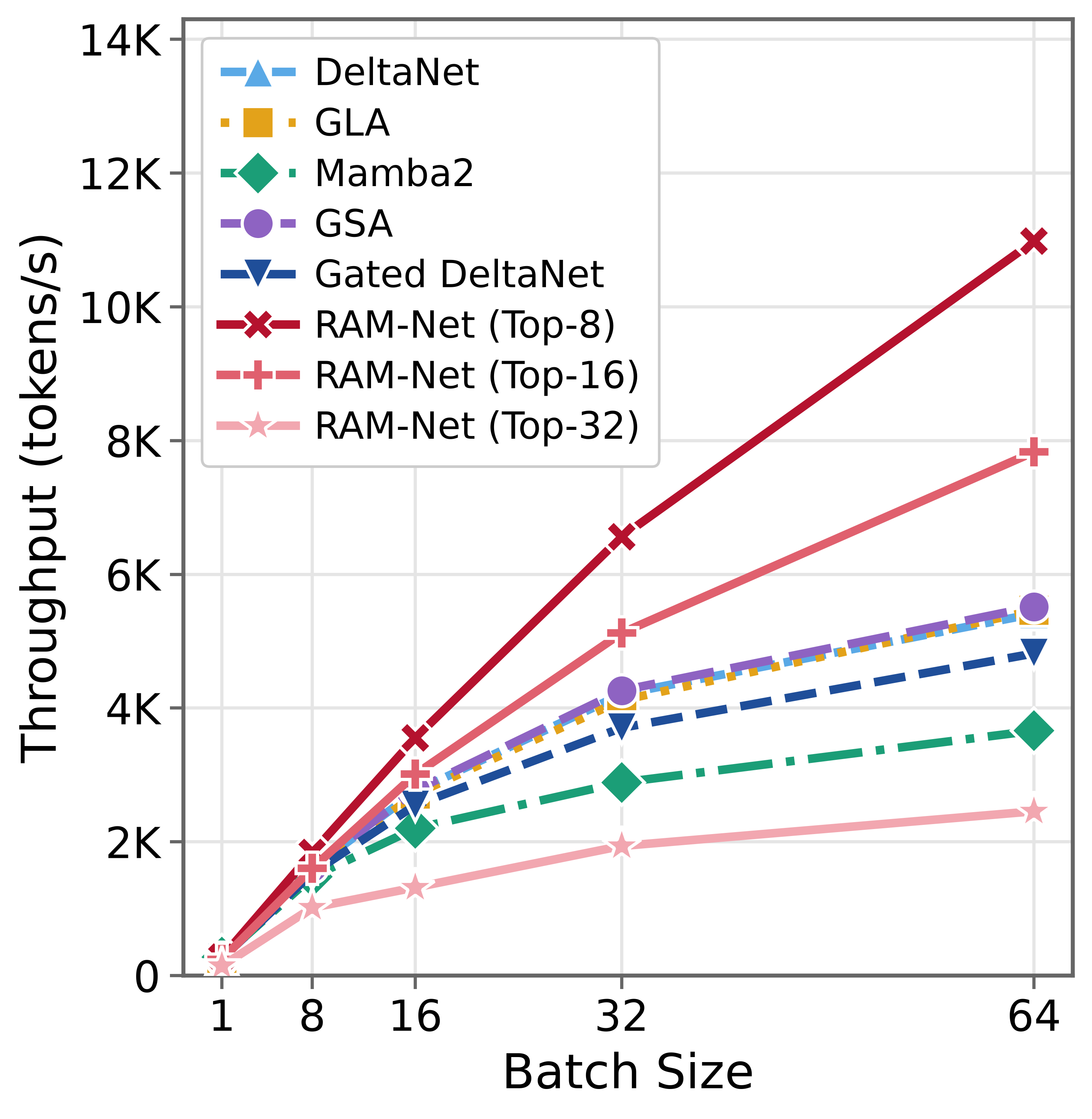}
    \caption{Decoding throughput on a single RTX 3060.}
    \vspace{-0.6cm}
    \label{fig:tps}
    \vspace{-0.3cm}
\end{wrapfigure}
We benchmark autoregressive decoding throughput of the 340M models on a single NVIDIA RTX 3060 as shown in Fig.~\ref{fig:tps}.
Each decoding step is captured in a CUDA graph to exclude Python and kernel-launch overhead.
At batch size 1, decoding is bound by reading the model weights, and all models run at similar speed.
As the batch grows, reading and updating the recurrent state becomes the bottleneck.
Since RAM-Net Top-8 accesses the least state per step, it achieves $2.0\times$ the throughput of the fastest baseline at batch size 64, although its total state is the largest among the recurrent models in Fig.~\ref{fig:tps}.
This shows that RAM-Net decouples decoding speed from total state size.
Its large total state supports the strong retrieval in Table~\ref{tab:niah}, while its sparse access keeps decoding fast.
The remaining trade-off is set by $K$, as a larger $K$ improves accuracy but lowers throughput.

\begin{table*}[t]
\centering
\caption{Zero-shot performance on recall-intensive tasks. FDA, SWDE, and NQ are evaluated at context lengths 512 and 1K; the remaining tasks use the full document.}
\label{tab:jrt}
\scriptsize
\resizebox{\textwidth}{!}{
\begin{tabular}{l|ccccccccc|c}
\toprule
\textbf{Model} &
\shortstack{\textbf{FDA}\\\footnotesize 512 $\uparrow$} &
\shortstack{\textbf{FDA}\\\footnotesize 1K $\uparrow$} &
\shortstack{\textbf{SWDE}\\\footnotesize 512 $\uparrow$} &
\shortstack{\textbf{SWDE}\\\footnotesize 1K $\uparrow$} &
\shortstack{\textbf{NQ}\\\footnotesize 512 $\uparrow$} &
\shortstack{\textbf{NQ}\\\footnotesize 1K $\uparrow$} &
\shortstack{\textbf{SQuADv2}\\\footnotesize full $\uparrow$} &
\shortstack{\textbf{TriviaQA}\\\footnotesize full $\uparrow$} &
\shortstack{\textbf{DROP}\\\footnotesize full $\uparrow$} &
\textbf{Avg.} \\
\midrule
\multicolumn{11}{l}{\textit{340M}} \\
Transformer++ & \tb{50.5} & \tb{37.8} & \tb{44.2} & \tb{37.8} & \tb{23.5} & \tb{24.8} & \tb{35.3} & 44.7 & 17.0 & \tb{35.1} \\
Gated DeltaNet & 22.1 & 15.1 & 37.8 & 17.0 & 21.8 & \ul{22.7} & \ul{33.0} & \ul{47.7} & 17.7 & 26.1 \\
DeltaNet & 22.4 & 16.1 & 32.0 & 9.5 & 20.6 & 20.2 & 27.7 & 37.3 & \tb{21.7} & 23.1 \\
Mamba2 & 29.8 & 22.1 & 36.7 & 18.7 & \ul{22.3} & 20.6 & 32.3 & 44.3 & 15.3 & 26.9 \\
GLA & 11.0 & 7.4 & 25.5 & 11.9 & 17.6 & 18.9 & 25.7 & 40.7 & 15.7 & 19.4 \\
GSA & 7.4 & 5.4 & 25.2 & 9.9 & 16.0 & 15.5 & 25.0 & 40.0 & 17.7 & 18.0 \\
MoM & 14.7 & 9.0 & 30.6 & 18.7 & 18.9 & 16.8 & 28.0 & 42.0 & 16.3 & 21.7 \\
\rowcolor{ramrow}
RAM-Net & & & & & & & & & & \\
\rowcolor{ramrow}
\quad Top-8 & 29.8 & 20.4 & 30.3 & 16.7 & 17.6 & 18.1 & 24.3 & 34.7 & 16.7 & 23.2 \\
\rowcolor{ramrow}
\quad Top-16 & 38.8 & 27.8 & 40.5 & 19.4 & 21.4 & 17.2 & 29.7 & 38.3 & 18.0 & 27.9 \\
\rowcolor{ramrow}
\quad Top-32 & \ul{46.8} & \ul{36.8} & \ul{42.5} & \ul{24.8} & 21.8 & 21.4 & 26.7 & \tb{48.3} & \ul{18.7} & \ul{32.0} \\
\midrule
\multicolumn{11}{l}{\textit{1.3B}} \\
Transformer++ & \tb{79.9} & \tb{70.6} & \tb{48.6} & \tb{48.6} & \tb{36.1} & \tb{35.7} & \tb{49.7} & 56.3 & \ul{22.3} & \tb{49.8} \\
Gated DeltaNet & 54.2 & 41.8 & \ul{43.9} & \ul{40.8} & \ul{34.9} & 32.4 & \ul{39.7} & \tb{61.3} & \ul{22.3} & 41.3 \\
Mamba2 & 48.5 & 34.4 & 39.5 & 29.9 & 34.5 & \ul{34.5} & 38.0 & 57.0 & 21.3 & 37.5 \\
\rowcolor{ramrow}
RAM-Net & & & & & & & & & & \\
\rowcolor{ramrow}
\quad Top-8 & 65.9 & 56.5 & 37.1 & 36.4 & 27.7 & 26.5 & 34.7 & \ul{57.3} & \tb{23.3} & 40.6 \\
\rowcolor{ramrow}
\quad Top-16 & 60.5 & 49.5 & 41.2 & 36.7 & 29.8 & 31.1 & 30.0 & 56.0 & 16.3 & 39.0 \\
\rowcolor{ramrow}
\quad Top-32 & \ul{68.6} & \ul{58.9} & 42.9 & 36.4 & \tb{36.1} & 29.0 & 34.7 & 56.3 & 21.3 & \ul{42.7} \\
\bottomrule
\end{tabular}
}
\vspace{-0.3cm}
\end{table*}

\section{Related Work} 
\paragraph{Linear Recurrent Models and State Expansion.}
Linear recurrent models replace the growing KV cache of full attention with a fixed-size recurrent state~\citep{katharopoulos2020transformers, sun2023retentive}.
Early variants, such as Linear Transformer~\citep{katharopoulos2020transformers} and RetNet~\citep{sun2023retentive}, aggregate history through kernel feature maps or fixed decay rules that do not depend on the input.
Subsequent models make the state update data-dependent.
Specifically, Mamba2~\citep{dao2024transformers}, GLA~\citep{yang2024gla}, and HGRN2~\citep{qin2024hgrn2gatedlinearrnns} use input-driven gating or decay, whereas DeltaNet~\citep{schlag2021linear}, Gated DeltaNet~\citep{yang2024gated}, RWKV-7~\citep{peng2025rwkv}, and Kimi Linear~\citep{kimiteam2025kimilinear} edit the state through delta-rule error correction.
TTT~\citep{sun2024learning} further views the state as fast weights trained online on a self-supervised loss.

Another line of work targets the limited capacity of fixed recurrent states.
MoM~\citep{du2025mom} routes each token to a few of several memory states, and SSE~\citep{pan2025scaling} similarly selects a few partitions of an expanded state, so that different tokens update different regions.
However, each selected region remains a dense matrix that is updated and read as a whole.
In contrast, GSA~\citep{zhang2024gsa} organizes the state into finer units called slots, but updates all slots with gated decay and reads them through a softmax at every step.
Instead, RAM-Net makes each individual slot the unit of selective access, as each key or query addresses only a few of many slots.
The per-step access therefore stays small while the number of slots can grow, and interference is confined to co-addressed slots, leaving all other slots unchanged.

\paragraph{Sparse Memory Addressing.}
Beyond recurrent states, discrete content-based addresses have also been used to retrieve from large sets of key-value pairs.
Transformer-VQ~\citep{lingle2023transformer} quantizes each key to a codeword and aggregates values under the codewords, but each query still reads all codewords densely.
Conversely, Product-Key Memory (PKM)~\citep{lample2019large} retrieves only the top-$k$ of millions of slots through a product search over two sub-key sets.
However, PKM slots are slow weights fixed at inference, so they store dataset-level knowledge rather than the current sequence.
FwPKM~\citep{zhao2026fast} extends PKM into a fast-weight memory that updates the selected slots online by gradient descent on a reconstruction loss.
Nevertheless, FwPKM is an auxiliary module placed before an attention or linear attention layer.
It updates its memory only once per chunk and has no explicit forgetting.
In contrast, RAM-Net is itself an attention layer whose recurrent state is read and written through sparse addresses.
Its Address Decoder factorizes each address into $U$ digits and produces the write and read weights at every step.
GSU then updates the few written slots with learned forgetting.

\paragraph{Sparse Attention.}
Sparse attention reduces the cost of full attention by restricting each query to a subset of past tokens.
Early methods use predefined sparse patterns, such as the fixed and strided factorizations in Sparse Transformer~\citep{child2019generating}, the sliding-window and global attention in Longformer~\citep{beltagy2020longformer}, the global-local attention in ETC~\citep{ainslie2020etc}, and the local-global-random pattern in BigBird~\citep{zaheer2020big}.
Subsequent methods make the sparse pattern content-dependent, for example through locality-sensitive hashing in Reformer~\citep{kitaev2020reformer}, online clustering in Routing Transformer~\citep{roy2021efficient}, differentiable sorting in Sparse Sinkhorn Attention~\citep{tay2020sparse}, or Top-$K$ selection in SparseK Attention~\citep{lou2024sparser}.
More recently, long-context systems select KV-cache blocks in a query-aware or hardware-aligned manner~\citep{tang2024quest, jiang2024minference, yuan2025native}, or retrieve relevant tokens through product quantization of cached keys~\citep{zhang2025pqcache}.
Despite these differences, these methods sparsify only the reads, as every token is still written into a KV cache that grows with sequence length.
In contrast, RAM-Net also sparsifies the writes, since each token writes only to its $K$ addressed slots in a fixed-size state.
As a result, its per-step cost stays constant in sequence length.

\section{Limitations and Future Work}
\label{sec:limit}

First, our experiments are limited to models with up to 1.3B parameters.
Whether the advantages of sparse addressing persist at larger scales remains to be verified.

Second, sparse addressing involves an inherent recall trade-off.
Restricting each read and write to a few slots reduces interference, but relevant content can be missed when the query is not co-addressed with it.
This limitation is most evident when the context contains diverse distractors over long ranges.
More flexible addressing, such as adaptive access, may mitigate this issue.

Third, GSU updates each addressed slot independently.
Richer update rules that coordinate co-addressed slots, such as delta-rule-style joint corrections~\citep{schlag2021linear, yang2024gated}, remain to be explored.

Finally, our implementation does not yet exploit the structure of the learned access patterns, in which a few slots are accessed far more frequently than others.
This skew suggests system-level optimizations, such as keeping frequently accessed slots in fast memory and storing the rest in lower-cost memory.

\section{Conclusion}
In this work, we introduced RAM-Net, a linear-time sequence model that replaces dense access to a shared recurrent state with sparse address-based read and write over a fixed slot array.
By making token interactions depend on overlap in addressed slots, RAM-Net restricts retrieval interference to a small subset of the state, while its per-token state access remains independent of the total number of slots and its addressing cost grows only logarithmically with it.
Empirically, RAM-Net shows clear advantages on fine-grained retrieval benchmarks over recurrent baselines, and remains competitive on standard language modeling and commonsense reasoning tasks.
These results suggest that explicit sparse addressing is a promising paradigm for building fixed-state sequence models with improved long-range recall.

\begin{ack}
This work was supported by the Research Grants Council (RGC) of Hong Kong
under Grant CUHK 14212323 and CUHK R1018-25.
The authors declare no competing interests.
\end{ack}

{
\small
\bibliography{ref}
\bibliographystyle{plainnat}
}

\newpage
\appendix

\section{Training Details}
\label{sec:training_details}
\paragraph{Model and Training Configuration.}
Tables~\ref{tab:model_config} and~\ref{tab:training_config} summarize the RAM-Net architecture and the training setup at the 340M and 1.3B scales.
All baselines follow the default configurations of the Flash-Linear-Attention library \citep{yang2024fla, yang2025flame} and are trained with the same setup.

In each layer, some heads use CAPE and the remaining heads use NoPE.
The number of CAPE heads decreases stage-wise with depth, starting from all heads in the first four layers.
Writing $h^{\times l}$ for $l$ consecutive layers with $h$ CAPE heads each, the 340M model uses
$\{16^{\times4}, 8^{\times4}, 4^{\times4}, 2^{\times4}, 1^{\times2}, 0^{\times7}\}$,
and the 1.3B model uses
$\{32^{\times4}, 16^{\times4}, 8^{\times4}, 4^{\times4}, 2^{\times9}\}$.

\begin{table}[ht]
\centering
\caption{Model configurations.}
\label{tab:model_config}
\small
\setlength{\tabcolsep}{10pt}
\renewcommand{\arraystretch}{1.1}
\begin{tabular}{l|cc}
\toprule
Params. & \textbf{340M} & \textbf{1.3B} \\
\midrule
Layers & \multicolumn{2}{c}{25} \\
Hidden size & 1,024 & 2,048 \\
MLP intermediate size & 2,816 & 5,632 \\
Heads & 16 & 32 \\
Value channels per head & \multicolumn{2}{c}{64} \\
\bottomrule
\end{tabular}
\end{table}

\begin{table}[ht]
\centering
\caption{Training configuration.}
\label{tab:training_config}
\small
\setlength{\tabcolsep}{6pt}
\renewcommand{\arraystretch}{1.05}
\begin{tabular}{l|l|c c}
\toprule
\textbf{Category} & \textbf{Setting} & \textbf{340M params.} & \textbf{1.3B params.} \\
\midrule
\multirow{2}{*}{Data}
    & Dataset                & \multicolumn{2}{c}{FineWeb-Edu \citep{lozhkov2024fineweb-edu}} \\
    & Training budget        & 10B tokens & 100B tokens \\
\midrule
\multirow{3}{*}{Tokenizer}
    & Tokenizer              & \multicolumn{2}{c}{Llama 2 \citep{touvron2023llama}} \\
    & Vocabulary size        & \multicolumn{2}{c}{32k} \\
    & Context window         & \multicolumn{2}{c}{4{,}096} \\
\midrule
\multirow{3}{*}{Optimization}
    & Optimizer              & \multicolumn{2}{c}{AdamW \citep{loshchilov2019decoupledweightdecayregularization}} \\
    & Weight decay           & \multicolumn{2}{c}{0.1} \\
    & Gradient clipping      & \multicolumn{2}{c}{1.0} \\
\midrule
\multirow{4}{*}{LR schedule}
    & Schedule               & \multicolumn{2}{c}{Cosine decay \citep{loshchilov2017sgdrstochasticgradientdescent}} \\
    & Linear warm-up         & 500M tokens          & 1B tokens \\
    & Peak LR                & $1.0\times 10^{-3}$  & $4.0\times 10^{-4}$ \\
    & Min LR                 & $1.0\times 10^{-4}$  & $4.0\times 10^{-5}$ \\
\midrule
\multirow{3}{*}{Batch}
    & Per-device batch       & \multicolumn{2}{c}{65.5K tokens} \\
    & Global batch size      & \multicolumn{2}{c}{0.5M tokens} \\
    & Devices             & \multicolumn{2}{c}{NVIDIA H200} \\
\bottomrule
\end{tabular}
\end{table}

\paragraph{Learnable Address Temperature.}
In addition to the fixed temperature $\tau$ in \S\ref{sec:addr_decode}, each head learns a log-scale correction for every digit of its write and read addresses.
Specifically, the digit logits of the key are scaled as $\exp(\beta_w^{(u)})\,\mathbf{k}_t^{(u)}/\tau$, and those of the query as $\exp(\beta_r^{(u)})\,\mathbf{q}_t^{(u)}/\tau$, where $\beta_w^{(u)}$ and $\beta_r^{(u)}$ are learnable scalars initialized to zero.
The effective temperature is thus $\tau e^{-\beta}$, and the exponential parameterization keeps it positive.
This allows each head to sharpen or smooth its addresses during training, starting from the base temperature $\tau=\sqrt U$.

\begin{figure}[t]
    \centering
    \begin{minipage}[t]{0.45\linewidth}
        \centering
        \includegraphics[width=0.9\linewidth]{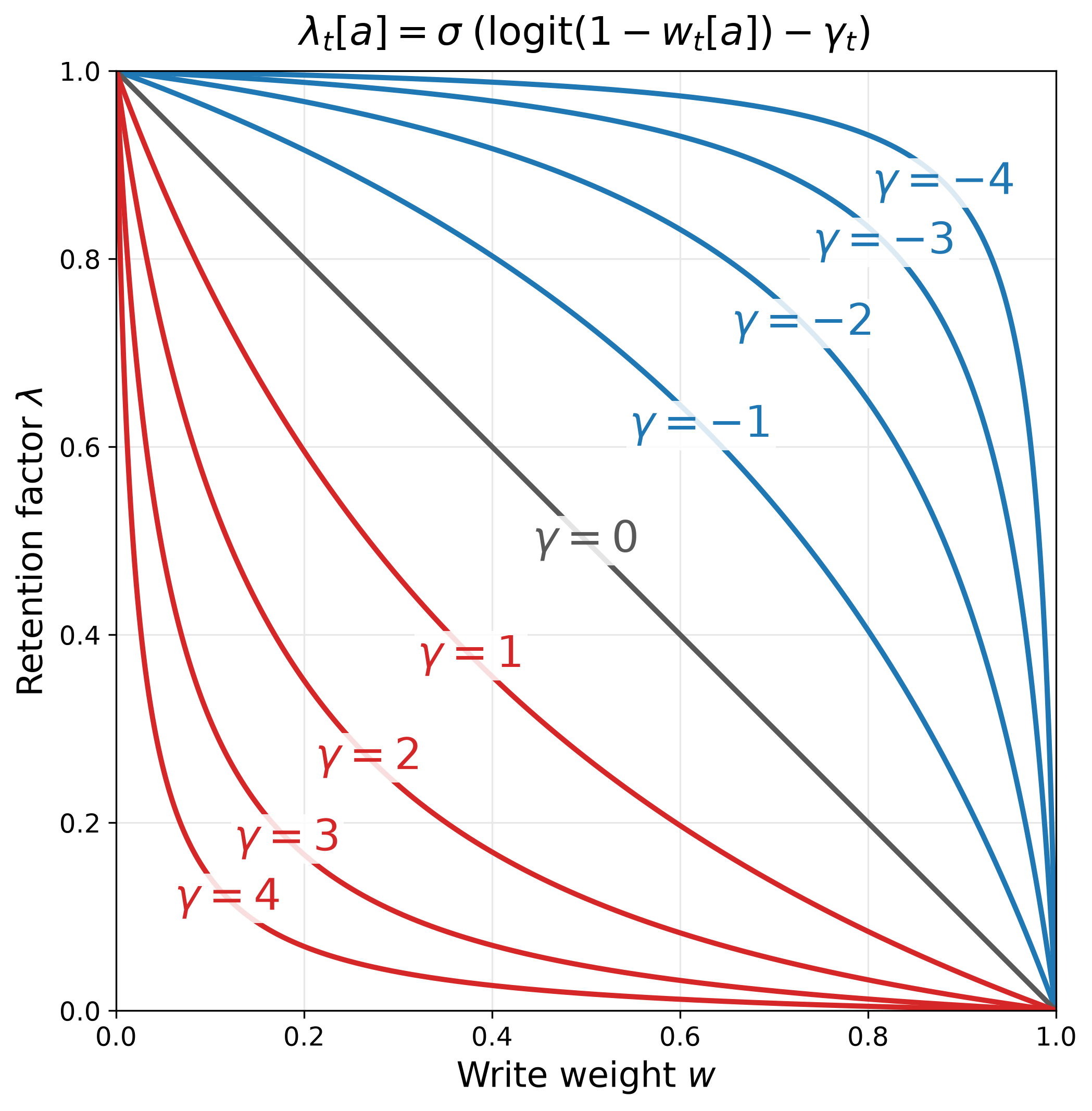}
        \caption{Retention factor $\lambda_t[a]$ as a function of the write weight $w_t[a]$ for different values of the forgetting modulator $\gamma_t$.
        }
        \label{fig:gamma}
    \end{minipage}
    \hfill
    \begin{minipage}[t]{0.53\linewidth}
        \centering
        \includegraphics[width=0.9\linewidth]{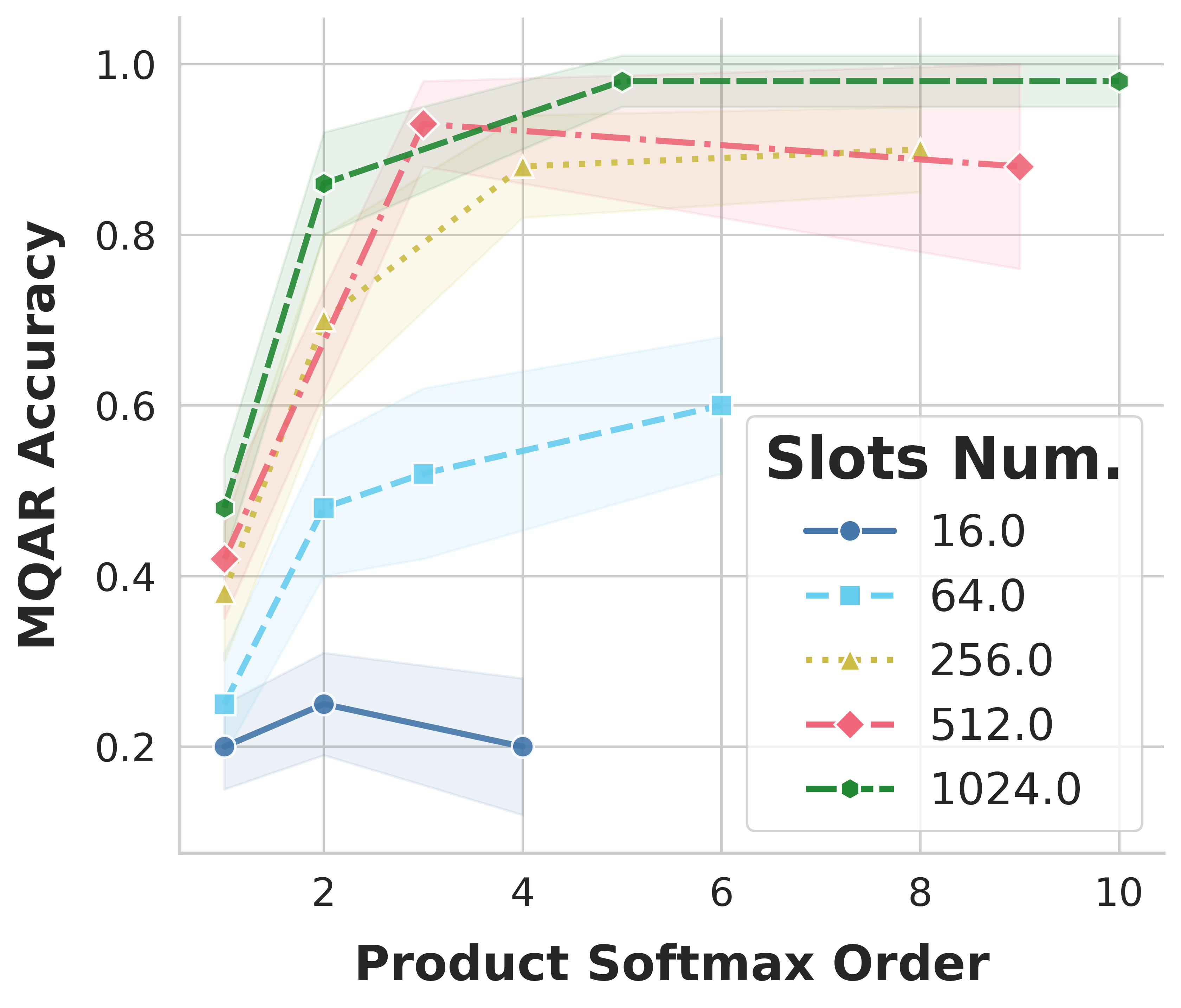}
       \caption{MQAR accuracy versus the number of address digits $U$ for different numbers of slots $|\mathcal{A}|$, where increasing $U$ improves accuracy.}
        \label{fig:ablation_u}
    \end{minipage}
\end{figure}

\paragraph{Effect of the Forgetting Modulator.}
Fig.~\ref{fig:gamma} shows how the forgetting modulator $\gamma_t$ reshapes the retention factor $\lambda_t[a]$ as a function of the write weight $w_t[a]$.
At $\gamma_t=0$, the retention factor equals $1-w_t[a]$, so the stronger a write is, the less of its previous mass the slot keeps.
A positive $\gamma_t$ lowers this curve, so the slot discards more of its accumulated mass and adopts new values faster.
Conversely, a negative $\gamma_t$ raises the curve, so the slot retains more of its mass and averages over a longer history.
All curves share the same endpoints regardless of $\gamma_t$.
At $w_t[a]=0$, $\lambda_t[a]=1$, so the modulator leaves unaddressed slots unchanged and preserves the exact sparsity property (\S\ref{sec:gsu}) of GSU.
At $w_t[a]=1$, $\lambda_t[a]=0$, so a slot written with full weight is always overwritten.

\paragraph{Bias Initialization.}
Sparse addressing and gated updates are sensitive to initialization, so we initialize three groups of biases with fixed values.
First, the output gate bias is set to $3$, so that the gate starts nearly open at $\sigma(3)\approx0.95$ and does not suppress the layer output, which lets gradients reach the addressing and update parameters from the start of training.
Second, we add a fixed bias to the address logits so that slots start with distinct preferences rather than being symmetric.
Within digit $u$, this bias increases linearly across the $d_p$ choices from $-\alpha_u$ to $\alpha_u$, and its magnitude $\alpha_u$ grows linearly across digits from $0$ to $r$.
The key and query projections share this bias, and we use $r = 1.0$ for the 340M models and $r = 1.5$ for the 1.3B models. 
Setting $r = 0$ removes the bias and markedly degrades retrieval at 340M (Table~\ref{tab:ablation}).

\section{Ablation Studies}
\label{sec:ablation}
\subsection{Ablation on Product Softmax and Slot Count} \label{sec:abla_prod}
We conduct an ablation study on MQAR \citep{arora2024zoology} to examine how the Address Decoder affects retrieval capability.
Specifically, we vary the Product Softmax order $U$ and the total slot count $\capa$, thereby separating the effect of Product Softmax from the effect of state size.

\paragraph{Order $U$ of Product Softmax.}
We first study the role of the Product Softmax order $U$.
When $U=1$, the address is produced by applying a single softmax over all $\capa$ slots followed by Top-$K$ truncation.
For $U>1$, the model applies softmax separately to $U$ low-dimensional sub-vectors, and their Kronecker product gives the probability of each slot before Top-$K$ selection.
As shown in Fig.~\ref{fig:ablation_u}, increasing $U$ substantially improves MQAR accuracy across most slot counts, with the gains most pronounced when the slot count is large.
This suggests that product softmax is important not only for efficient Top-$K$ decoding, but also for making large address spaces easier to optimize.

A likely reason is that, with $U=1$, the address is a single softmax over all $\capa$ slots, and after Top-$K$ truncation only the $K$ selected entries receive a direct learning signal at each step.
With $U>1$, each digit is shared by many slots, so the learning signal from a selected slot also reaches every slot that shares its digits.
This denser learning signal makes large address spaces easier to optimize, consistent with the larger gains at large slot counts in Fig.~\ref{fig:ablation_u}.

\paragraph{Slot Count $\capa$.}
We next study the effect of the total number of slots.
For a fixed or sufficiently large Product Softmax order, larger slot counts generally improve MQAR accuracy, with the best results obtained at $\capa=1024$.
This trend is consistent with the role of slots as addressable storage units: increasing $\capa$ enlarges the address space and reduces the chance that unrelated key-value pairs must share the same slots.

\begin{table*}[h]
\centering
\caption{Ablation study of RAM-Net design choices. The default setting uses $\capa=4^5$ slots, Top-$K=8$, and the CAPE setting in Appendix~\ref{sec:training_details}. All variants are trained with a 4K context window unless noted otherwise. Results are reported on language modeling, commonsense reasoning, and 4K S-NIAH retrieval tasks.}
\label{tab:ablation}
\resizebox{\textwidth}{!}{
\begin{tabular}{l|cccccccccccc|ccc}
\toprule
\textbf{Variant} &
\shortstack{\textbf{Wiki.}\\\footnotesize $\text{ppl}$ $\downarrow$} &
\shortstack{\textbf{LMB.}\\\footnotesize $\text{acc}$ $\uparrow$} &
\shortstack{\textbf{MMLU}\\\footnotesize $\text{acc}$ $\uparrow$} &
\shortstack{\textbf{ARC-e}\\\footnotesize $\text{acc}$ $\uparrow$} &
\shortstack{\textbf{ARC-c}\\\footnotesize $\text{acc}$ $\uparrow$} &
\shortstack{\textbf{OBQA}\\\footnotesize $\text{acc}_\text{n}$ $\uparrow$} &
\shortstack{\textbf{SciQ}\\\footnotesize $\text{acc}$ $\uparrow$} &
\shortstack{\textbf{boolQ}\\\footnotesize $\text{acc}$ $\uparrow$} &
\shortstack{\textbf{COPA}\\\footnotesize $\text{acc}$ $\uparrow$} &
\shortstack{\textbf{PIQA}\\\footnotesize $\text{acc}$ $\uparrow$} &
\shortstack{\textbf{Hella.}\\\footnotesize $\text{acc}$ $\uparrow$} &
\shortstack{\textbf{Wino.}\\\footnotesize $\text{acc}$ $\uparrow$} &
\shortstack{\textbf{S-NIAH-1}\\\footnotesize $\text{acc}$ $\uparrow$} &
\shortstack{\textbf{S-NIAH-2}\\\footnotesize $\text{acc}$ $\uparrow$} &
\shortstack{\textbf{S-NIAH-3}\\\footnotesize $\text{acc}$ $\uparrow$} \\
\midrule
\rowcolor{ramrow}
RAM-Net & & & & & & & & & & & & & & & \\
\rowcolor{ramrow}
\quad Top-8 (default) & 26.89 & 28.7 & 23.6 & 59.1 & 24.7 & 34.4 & 82.5 & 57.4 & 67.0 & 66.9 & 33.4 & 54.1 & 99.7 & 83.7 & 18.3 \\
\rowcolor{ramrow}
\quad Top-16 & 26.23 & 29.7 & 23.5 & 59.4 & 26.1 & 33.0 & 82.8 & 58.6 & 69.0 & 67.2 & 33.8 & 51.4 & 100.0 & 88.0 & 25.7 \\
\rowcolor{ramrow}
\quad Top-32 & 25.87 & 30.9 & 23.8 & 58.9 & 26.1 & 32.2 & 82.8 & 61.6 & 70.0 & 67.2 & 33.7 & 52.7 & 100.0 & 94.0 & 44.7 \\
\midrule
\midrule
\multicolumn{16}{l}{\textit{Ablations}} \\
\quad-- write Top-4, read Top-12 & 27.98 & 27.2 & 23.7 & 58.8 & 26.5 & 33.4 & 81.4 & 60.2 & 68.0 & 66.2 & 33.1 & 51.9 & 98.0 & 45.3 & 17.0 \\
\quad-- write Top-12, read Top-4 & 26.59 & 29.3 & 23.1 & 59.7 & 25.8 & 33.8 & 82.8 & 62.0 & 70.0 & 66.3 & 33.6 & 51.1 & 100.0 & 48.7 & 7.7 \\
\quad-- $\capa=4^4$ slots & 26.47 & 28.0 & 23.6 & 59.1 & 26.5 & 34.6 & 82.3 & 55.0 & 67.0 & 67.5 & 33.7 & 53.3 & 100.0 & 41.7 & 13.7 \\
\quad-- $\capa=4^6$ slots & 27.09 & 27.6 & 24.0 & 60.0 & 25.5 & 34.2 & 82.4 & 60.8 & 67.0 & 67.5 & 33.6 & 52.0 & 100.0 & 10.0 & 14.3 \\
\quad-- $\capa=4^6$ slots, 8K training & 26.90 & 26.8 & 23.9 & 58.6 & 25.1 & 33.6 & 81.0 & 54.9 & 67.0 & 66.4 & 33.1 & 51.6 & 100.0 & 34.0 & 7.0 \\
\midrule
\quad-- fixed $\gamma_t=0$ & 27.59 & 27.6 & 23.6 & 57.2 & 25.3 & 30.6 & 82.6 & 61.7 & 69.0 & 66.1 & 33.0 & 50.9 & 100.0 & 31.0 & 30.3 \\
\midrule
\quad-- w/o output RMSNorm & 28.87 & 26.3 & 23.8 & 57.9 & 24.1 & 33.0 & 82.0 & 59.6 & 67.0 & 66.6 & 32.6 & 53.1 & 99.7 & 16.3 & 12.3 \\
\quad-- w/o output gate & 27.56 & 28.6 & 23.6 & 57.4 & 25.0 & 32.0 & 81.5 & 59.8 & 70.0 & 65.6 & 32.8 & 51.6 & 100.0 & 50.3 & 25.0 \\
\midrule
\quad-- $r=0.0$ & 28.18 & 27.1 & 23.5 & 56.7 & 24.8 & 32.6 & 81.0 & 59.4 & 70.0 & 67.0 & 32.5 & 53.1 & 100.0 & 27.7 & 5.3 \\
\quad-- $r=1.5$ & 26.63 & 29.4 & 24.6 & 59.7 & 25.7 & 34.0 & 81.2 & 58.2 & 67.0 & 66.5 & 34.0 & 52.3 & 100.0 & 58.7 & 25.7 \\
\quad-- $\tau=1.5$ & 26.61 & 29.5 & 24.4 & 58.0 & 24.8 & 33.0 & 81.5 & 56.1 & 64.0 & 67.0 & 33.4 & 52.3 & 100.0 & 56.0 & 25.0 \\
\quad-- Top-$K$ renorm & 26.83 & 27.8 & 23.1 & 60.8 & 25.6 & 33.6 & 83.3 & 60.2 & 70.0 & 67.9 & 33.8 & 51.0 & 100.0 & 63.7 & 12.0 \\
\midrule
\quad-- NoPE only & 28.04 & 30.0 & 23.0 & 58.4 & 24.9 & 32.2 & 82.3 & 61.4 & 69.0 & 66.8 & 33.3 & 54.4 & 100.0 & 43.0 & 8.7 \\
\quad-- NoPE + conv & 27.13 & 30.2 & 22.8 & 58.9 & 24.7 & 32.8 & 81.0 & 55.4 & 70.0 & 66.0 & 33.2 & 50.6 & 100.0 & 16.3 & 1.0 \\
\quad-- CAPE + conv & 26.84 & 28.9 & 23.7 & 58.2 & 24.1 & 34.4 & 80.6 & 59.4 & 69.0 & 66.8 & 33.7 & 50.9 & 99.7 & 88.3 & 15.0 \\
\quad-- uniform CAPE & 26.90 & 28.0 & 23.8 & 57.7 & 24.0 & 33.2 & 81.4 & 61.6 & 66.0 & 67.0 & 33.3 & 51.1 & 99.7 & 22.7 & 10.0 \\
\quad-- 60\% CAPE & 27.12 & 27.9 & 24.6 & 57.8 & 25.7 & 32.6 & 79.4 & 57.9 & 70.0 & 66.4 & 33.4 & 50.1 & 100.0 & 16.3 & 6.0 \\
\midrule
\quad-- peak LR $10^{-2}$ & 30.86 & 25.0 & 23.4 & 56.9 & 25.3 & 32.8 & 81.7 & 57.6 & 69.0 & 64.5 & 32.4 & 52.2 & 100.0 & 14.7 & 9.0 \\
\quad-- peak LR $10^{-4}$ & 44.57 & 18.3 & 22.9 & 50.0 & 20.1 & 29.4 & 77.8 & 62.0 & 59.0 & 60.8 & 28.7 & 51.1 & 77.7 & 21.3 & 0.0 \\
\quad-- slot balance & 27.10 & 26.0 & 23.6 & 58.3 & 26.2 & 36.0 & 82.0 & 60.3 & 68.0 & 66.5 & 32.9 & 52.6 & 100.0 & 11.3 & 5.7 \\
\bottomrule
\end{tabular}
}
\end{table*}

\subsection{Ablation on Other Design Choices} \label{app:ablation}
Table~\ref{tab:ablation} evaluates the remaining design choices on language modeling and S-NIAH retrieval at the 340M scale.
The default configuration uses $|\mathcal{A}|=4^5$ slots, Top-$K=8$, $\tau=\sqrt{5}\approx 2.24$, $r=1.0$, the learned forgetting modulator $\gamma_t$, output RMSNorm and output gate, the hybrid CAPE configuration in Appendix~\ref{sec:training_details}, and a peak learning rate of $10^{-3}$.
We observe no loss spikes in any run in the table, and none of them requires a restart.
In the following, we discuss the results by design aspect.

\paragraph{Per-step state access.}
Accessing more slots consistently helps.
Increasing Top-$K$ from 8 to 32 lowers Wiki perplexity from $26.89$ to $25.87$ and raises S-NIAH-2/3 accuracy from $83.7/18.3$ to $94.0/44.7$.
At a fixed budget of 16 accessed slots, the symmetric split of eight writes and eight reads gives the best retrieval.
Writing 4 and reading 12 raises perplexity by $1.09$ and lowers S-NIAH-2 to $45.3$, whereas writing 12 and reading 4 slightly improves perplexity but lowers S-NIAH-2/3 to $48.7/7.7$.

\paragraph{Slot count.}
With Top-$K$ fixed, the default slot count $|\mathcal{A}|=4^5$ performs best on retrieval.
Reducing it to $4^4$ lowers S-NIAH-2 to $41.7$, consistent with the MQAR trend in Appendix~\ref{sec:abla_prod}, where fewer slots force more unrelated content to share the same slots.
However, increasing it to $4^6$ lowers S-NIAH-2 further to $10.0$ and raises perplexity by $0.20$.
Training this variant with an 8K context under the same token budget does not recover retrieval ($34.0/7.0$), so the degradation is not caused by an insufficient training context.
A likely reason is that, with $K$ fixed, enlarging the address space reduces the chance that a query is co-addressed with the slots holding the relevant content.
The slot count therefore balances interference at small $|\mathcal{A}|$ against missed reads at large $|\mathcal{A}|$.

\paragraph{Forgetting modulator.}
The learned modulator $\gamma_t$ matters for both language modeling and retrieval.
Fixing $\gamma_t=0$ reduces the retention factor to $1-w_t[a]$, so each slot forgets only in proportion to the current write weight.
This raises perplexity by $0.70$ and reduces S-NIAH-2 from $83.7$ to $31.0$.

\paragraph{Read-out.}
Output RMSNorm is essential for retrieval.
Removing it increases Wiki perplexity by $1.98$ and reduces S-NIAH-2 from $83.7$ to $16.3$.
This effect persists even though GSU already keeps each slot on the scale of the values, suggesting that output RMSNorm mainly stabilizes the read output, whose scale varies with the unnormalized read weights.
The output gate contributes less.
Removing it raises perplexity by $0.67$ and reduces S-NIAH-2 to $50.3$.

\paragraph{Addressing.}
The address bias, temperature, and normalization all affect retrieval.
Setting $r=0$ gives all slots the same initial addressing preference, increasing perplexity by $1.29$ and reducing S-NIAH-2/3 to $27.7/5.3$.
Sharpening the addresses, either by a larger bias ($r=1.5$) or a lower temperature ($\tau=1.5$), slightly improves perplexity but reduces S-NIAH-2 to $58.7$ and $56.0$, respectively.
Renormalizing the retained Top-$K$ weights to sum to one, instead of keeping them unnormalized as in the default, reduces S-NIAH-2 to $63.7$.

\paragraph{Positional encoding.}
CAPE is important for retrieval.
Using NoPE in all heads reduces S-NIAH-2/3 to $43.0/8.7$.
Adding a short convolution with kernel size 4 to NoPE (following Mamba2 and Gated DeltaNet) improves perplexity but further reduces retrieval to $16.3/1.0$, indicating that local token mixing cannot replace CAPE.
Adding the same convolution to the default configuration changes retrieval only marginally ($88.3/15.0$).
In contrast, changing the CAPE layout removes its gain.
Assigning CAPE to five heads in every layer (uniform CAPE) keeps the share at about 30\% but reduces S-NIAH-2 to $22.7$.
Raising the share to around 60\% with a similar decreasing schedule (60\% CAPE), namely $\{16^{\times 9}, 8^{\times 9}, 4^{\times 4}, 2^{\times 2}, 1^{\times 1}\}$, reduces it to $16.3$.
Both layouts fall below NoPE in all heads, suggesting that position-shifted addressing is beneficial in the early layers but displaces content-based addressing when extended to the later layers.

\paragraph{Optimization and regularization.}
The default peak learning rate follows the Flame setting~\citep{yang2025flame} shared by all baselines.
A larger rate of $10^{-2}$ degrades both perplexity and retrieval, while a smaller rate of $10^{-4}$ learns too slowly and lags far behind within the 10B-token budget.
We also test an MoE-style load-balancing loss~\citep{fedus2022switch} on the write addresses, which penalizes, for each head, the product of how often each slot is selected and the address probability it receives.
This loss reduces S-NIAH-2/3 to $11.3/5.7$, suggesting that the uneven slot usage learned by RAM-Net (Appendix~\ref{appendix:access_pattern}) is beneficial rather than a defect to be corrected.

\begin{figure}[h]
    \centering
    \includegraphics[width=0.8\linewidth]{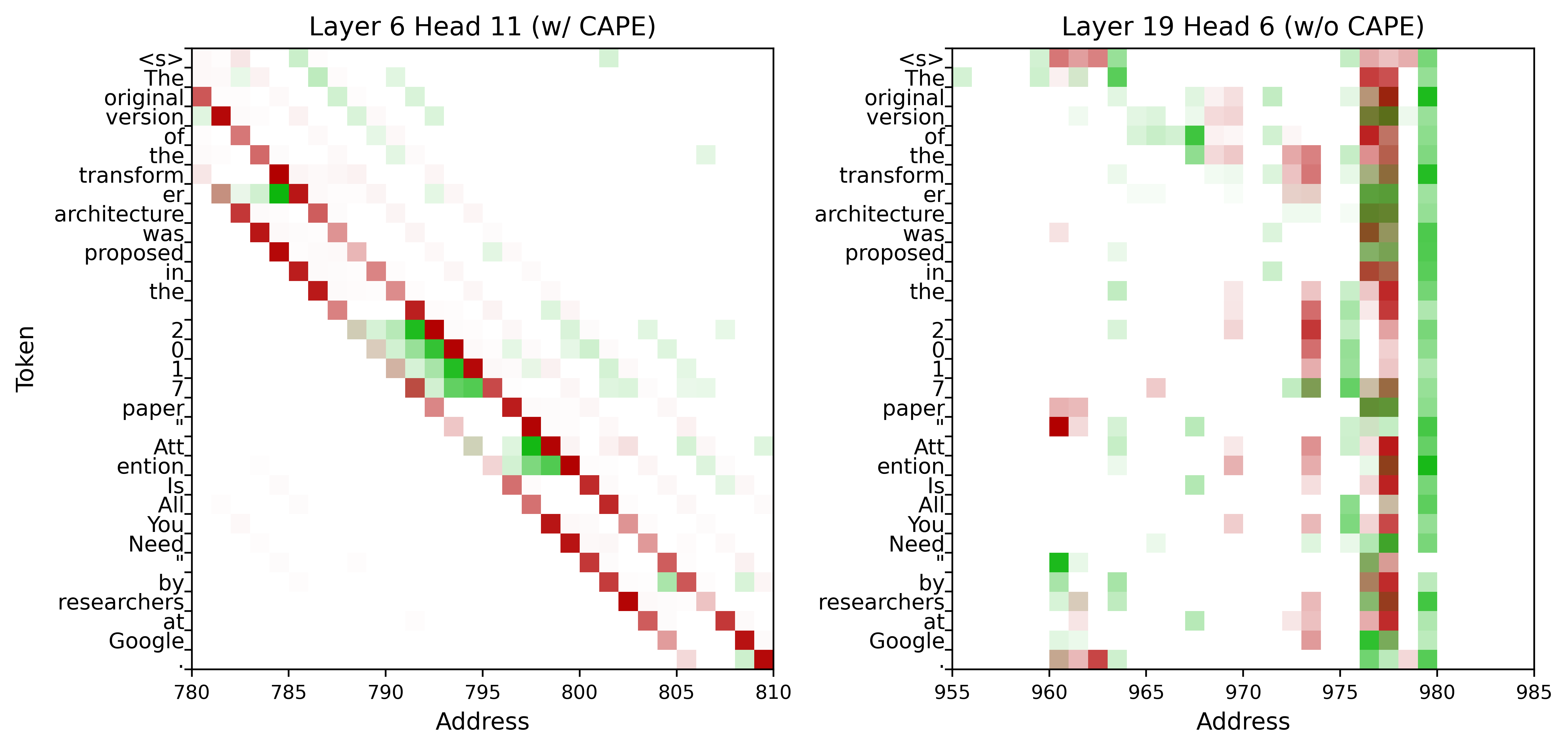}\\[-2pt]
    \begin{minipage}{0.8\linewidth}
        \centering
        \makebox[0.5\linewidth]{\footnotesize (a) CAPE head}%
        \makebox[0.5\linewidth]{\footnotesize (b) NoPE head}
    \end{minipage}
    \caption{Zoomed-in state access traces on the input
    \textit{``The original version of the transformer architecture was proposed in the 2017 paper `Attention Is All You Need' by researchers at Google.''}
    Each cell shows whether token $t$ (vertical axis) issues a {\color{ramnet-red}\textbf{write}} (red) or {\color{ramnet-green}\textbf{read}} (green) event at slot address (horizontal axis); color intensity encodes the corresponding write or read strength.
    (a) A CAPE head, in which the cyclic shift produces a clear diagonal trace.
    (b) A NoPE head, in which addresses are determined purely by content.}
    \label{fig:fine_trace}
\end{figure}

\section{State Access Trace Analysis}
\label{appendix:access_pattern}
\begin{figure}
    \centering
    \includegraphics[width=1\linewidth]{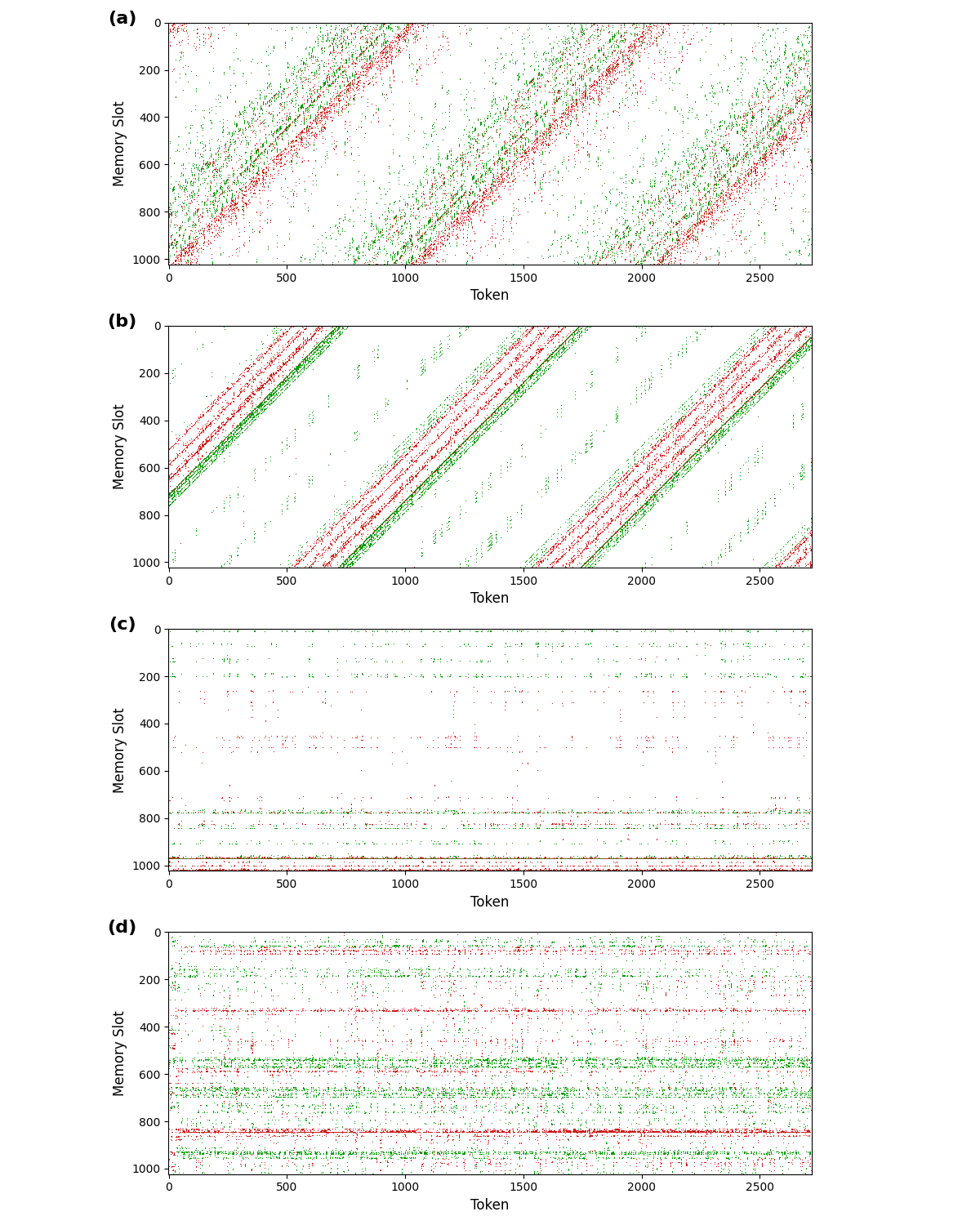}
    \caption{
    State access traces of four representative heads, showing {\color{ramnet-red}\textbf{write}} (red) and {\color{ramnet-green}\textbf{read}} (green) events at each slot address over tokens.
    CAPE heads (sub-figures (a) and (b)) produce diagonal traces, while NoPE heads (sub-figures (c) and (d)) show content-driven accesses.
    }
    \label{fig:coarse_trace}
\end{figure}

To better understand what RAM-Net learns under sparse addressing, we visualize the state access traces of two representative heads on a short factual sentence in Fig.~\ref{fig:fine_trace}, and the traces of four heads over a longer passage in Fig.~\ref{fig:coarse_trace}.
From these traces, we highlight three observations.

\paragraph{CAPE and NoPE heads exhibit complementary access styles.}
First, the CAPE heads (see Fig.~\ref{fig:fine_trace}a and Fig.~\ref{fig:coarse_trace}a,b) produce accesses concentrated along a diagonal.
This pattern directly reflects the cyclic shift in CAPE, which moves the slot index of a given content by one at each step, so that consecutive writes form a stride-one trace.
In contrast, the NoPE heads (Fig.~\ref{fig:fine_trace}b and Fig.~\ref{fig:coarse_trace}c,d) show no such drift, as their accesses depend only on token content and therefore fall on the same set of slots regardless of token position.
This divergence suggests a functional differentiation between the two head types, with CAPE heads organizing slots by position and NoPE heads relying on content for direct lookup.


\paragraph{Slot allocation within NoPE heads splits into shared and selective groups.}
Beyond the head-level distinction above, NoPE heads further partition their slots by usage frequency.
Specifically, a small group of slots receives reads and writes from nearly every token (e.g., addresses 977--980 in Fig.~\ref{fig:fine_trace}b, and the dense horizontal bands in Fig.~\ref{fig:coarse_trace}d), while the remaining slots are accessed only by a small subset of tokens with matching content.
Such an emergence of always-on slots is consistent with the ``attention sink'' phenomenon reported in full attention~\citep{xiao2024efficient}, where a few token positions absorb a large share of attention weight across queries.
Notably, attention sinks anchor at fixed token positions, whereas RAM-Net's shared slots emerge from learned address patterns.
Overall, these observations indicate that sparse addressing is not used uniformly.
Much like an attention sink, the model spontaneously sets aside a small fraction of its slots as sequence-level shared storage and reserves the rest for content-specific storage.

\paragraph{Sparse addressing links related tokens through overlapping slots.}
Finally, the CAPE head in Fig.~\ref{fig:fine_trace}a suggests that related tokens tend to access overlapping slot regions.
In the trace, such a link appears as a red write and a green read from different tokens falling in the same slot column.
We highlight three representative cases.
(i) \emph{Adjacent subword tokens.}
The second sub-token of \textit{transformer} (\textit{``er''}) reads from the slot just written by its first sub-token (\textit{``transform''}), and the same holds for \textit{``Att''} and \textit{``ention''}.
This suggests that the model learns to reconnect words split by the tokenizer.
(ii) \emph{Consecutive tokens forming one entity.}
The four digit tokens \textit{``2''}, \textit{``0''}, \textit{``1''}, and \textit{``7''} concentrate their reads and writes in a common slot region more densely than the surrounding tokens, suggesting that the model groups tokens that jointly denote a single entity.
(iii) \emph{Non-adjacent paired tokens.}
The opening and closing quotation marks of the cited title access the same slot region although several tokens separate them, so the closing mark may retrieve content written between the pair.
Since this is a CAPE head, these links depend jointly on token content and relative position, consistent with the design of CAPE in \S\ref{sec:cape}.


\newcommand{\rdd}{\textsuperscript{\textcolor{red}{$\dagger$}}}
\newcommand{\bdd}{\textsuperscript{\textcolor{blue}{$\dagger$}}}

\clearpage
\section*{NeurIPS Paper Checklist}

\begin{enumerate}

\item {\bf Claims}
    \item[] Question: Do the main claims made in the abstract and introduction accurately reflect the paper's contributions and scope?
    \item[] Answer: \answerYes{} 
    \item[] Justification: We have elucidated our contributions and scope in the abstract and introduction.

\item {\bf Limitations}
    \item[] Question: Does the paper discuss the limitations of the work performed by the authors?
    \item[] Answer: \answerYes{} 
    \item[] Justification: The limitations of this paper are discussed in Section \ref{sec:limit}.

\item {\bf Theory assumptions and proofs}
    \item[] Question: For each theoretical result, does the paper provide the full set of assumptions and a complete (and correct) proof?
    \item[] Answer: \answerNA{} 
    \item[] Justification: As our work is primarily empirical, it does not involve theoretical claims that require formal proofs.

    \item {\bf Experimental result reproducibility}
    \item[] Question: Does the paper fully disclose all the information needed to reproduce the main experimental results of the paper to the extent that it affects the main claims and/or conclusions of the paper (regardless of whether the code and data are provided or not)?
    \item[] Answer: \answerYes{} 
    \item[] Justification: The detailed training and evaluation settings are provided in Section \ref{sec:exp} and Appendix \ref{sec:training_details}.

\item {\bf Open access to data and code}
    \item[] Question: Does the paper provide open access to the data and code, with sufficient instructions to faithfully reproduce the main experimental results, as described in supplemental material?
    \item[] Answer: \answerYes{} 
    \item[] Justification: We provide the PyTorch and CUDA implementations in the Supplementary Material. All datasets used in this paper are reported in Section \ref{sec:exp} and publicly available.

\item {\bf Experimental setting/details}
    \item[] Question: Does the paper specify all the training and test details (e.g., data splits, hyperparameters, how they were chosen, type of optimizer) necessary to understand the results?
    \item[] Answer: \answerYes{} 
    \item[] Justification: We have detailed all the training and evaluation settings in Section \ref{sec:exp} and Appendix \ref{sec:training_details}.

\item {\bf Experiment statistical significance}
    \item[] Question: Does the paper report error bars suitably and correctly defined or other appropriate information about the statistical significance of the experiments?
    \item[] Answer: \answerNo{} 
    \item[] Justification: We do not have enough resources to obtain error bars, as LLM pre-training consumes a significant amount of computational power, making it impractical to conduct multiple experiments.

\item {\bf Experiments compute resources}
    \item[] Question: For each experiment, does the paper provide sufficient information on the computer resources (type of compute workers, memory, time of execution) needed to reproduce the experiments?
    \item[] Answer: \answerYes{} 
    \item[] Justification:  We provide the GPU type used for conducting our experiments in Appendix \ref{sec:training_details}.
    
\item {\bf Code of ethics}
    \item[] Question: Does the research conducted in the paper conform, in every respect, with the NeurIPS Code of Ethics \url{https://neurips.cc/public/EthicsGuidelines}?
    \item[] Answer: \answerYes{} 
    \item[] Justification: This work follows the NeurIPS Code of Ethics.

\item {\bf Broader impacts}
    \item[] Question: Does the paper discuss both potential positive societal impacts and negative societal impacts of the work performed?
    \item[] Answer: \answerNA{} 
    \item[] Justification: To our knowledge, this work poses no direct societal risks.
    
\item {\bf Safeguards}
    \item[] Question: Does the paper describe safeguards that have been put in place for responsible release of data or models that have a high risk for misuse (e.g., pre-trained language models, image generators, or scraped datasets)?
    \item[] Answer: \answerNA{} 
    \item[] Justification: To our knowledge, this work poses no such risks.

\item {\bf Licenses for existing assets}
    \item[] Question: Are the creators or original owners of assets (e.g., code, data, models), used in the paper, properly credited and are the license and terms of use explicitly mentioned and properly respected?
    \item[] Answer: \answerYes{} 
    \item[] Justification: All of the datasets we use are publicly available, and we have properly cited all the training and evaluation datasets we used in Section \ref{sec:exp}.

\item {\bf New assets}
    \item[] Question: Are new assets introduced in the paper well documented and is the documentation provided alongside the assets?
    \item[] Answer: \answerYes{} 
    \item[] Justification: Related codes of this work are released with suitable license.

\item {\bf Crowdsourcing and research with human subjects}
    \item[] Question: For crowdsourcing experiments and research with human subjects, does the paper include the full text of instructions given to participants and screenshots, if applicable, as well as details about compensation (if any)? 
    \item[] Answer: \answerNA{} 
    \item[] Justification: This work does not involve crowdsourcing nor research with human subjects.

\item {\bf Institutional review board (IRB) approvals or equivalent for research with human subjects}
    \item[] Question: Does the paper describe potential risks incurred by study participants, whether such risks were disclosed to the subjects, and whether Institutional Review Board (IRB) approvals (or an equivalent approval/review based on the requirements of your country or institution) were obtained?
    \item[] Answer: \answerNA{} 
    \item[] Justification: This work does not involve crowdsourcing nor research with human subjects.

\item {\bf Declaration of LLM usage}
    \item[] Question: Does the paper describe the usage of LLMs if it is an important, original, or non-standard component of the core methods in this research? Note that if the LLM is used only for writing, editing, or formatting purposes and does \emph{not} impact the core methodology, scientific rigor, or originality of the research, declaration is not required.
    \item[] Answer: \answerNA{} 
    \item[] Justification: LLM is used only for writing.

\end{enumerate}
\end{document}